\documentclass[10pt,letterpaper,twocolumn]{article}

\usepackage{times}
\usepackage{amsmath,amssymb,amsthm}
\usepackage{booktabs}
\usepackage{array}
\usepackage{enumitem}
\usepackage{url}
\usepackage{xcolor}
\usepackage{xspace}
\usepackage{hyperref}
\usepackage{xurl}
\usepackage[switch]{lineno}
\usepackage{fancyhdr}
\usepackage{caption}
\usepackage{graphicx}
\usepackage{pgfplots}
\pgfplotsset{compat=1.18}
\usetikzlibrary{shapes.geometric,arrows.meta,positioning,fit,calc,backgrounds}
\usepackage{listings}
\lstdefinestyle{lean}{%
  aboveskip=0pt,belowskip=0pt,
  basicstyle=\ttfamily\scriptsize,
  breaklines=true,breakindent=1.4em,
  columns=fullflexible,keepspaces=true,
  showstringspaces=false,extendedchars=true,
  literate=
    {ℕ}{{$\mathbb{N}$}}1 {ℝ}{{$\mathbb{R}$}}1
    {∀}{{$\forall$}}1 {∃}{{$\exists$}}1
    {∧}{{$\land$}}1 {∨}{{$\lor$}}1
    {≤}{{$\leq$}}1 {≥}{{$\geq$}}1 {→}{{$\to$}}1
    {ᶜ}{{$^{\mathsf c}$}}1 {↑}{{$\uparrow$}}1
}

\definecolor{cdef}{HTML}{9ECAE1}      
\definecolor{cthm}{HTML}{FDBE85}      
\definecolor{ctotal}{HTML}{7A4A12}    
\definecolor{clane}{HTML}{3B6EA5}     
\definecolor{clean}{HTML}{222222}     
\definecolor{cgreen}{HTML}{3F7A35}    
\definecolor{cgreenfill}{HTML}{C9E0C0}
\definecolor{credfill}{HTML}{E2706A}  
\definecolor{cred}{HTML}{A01F1F}      
\definecolor{cphase}{HTML}{777777}    
\definecolor{cdefnode}{HTML}{2B7BBA}  
\definecolor{cthmnode}{HTML}{E6863C}  
\definecolor{ctgt}{HTML}{C0182B}      
\definecolor{cdepedge}{HTML}{C8C8C8}  
\definecolor{ctgtedge}{HTML}{D24A3A}  
\definecolor{ctgtlabeltext}{HTML}{7A0010}

\pgfdeclarelayer{bg}\pgfsetlayers{background,bg,main}
\tikzset{
  defnode/.style={rectangle,rounded corners=0.5pt,fill=cdefnode,
                  draw=cdefnode!55!black,line width=0.4pt,
                  minimum size=4pt,inner sep=0pt},
  thmnode/.style={circle,fill=cthmnode,draw=cthmnode!55!black,line width=0.4pt,
                  minimum size=4pt,inner sep=0pt},
  closurenode/.style={rectangle,rounded corners=0.6pt,fill=ctgt,
                  draw=ctgtlabeltext,line width=0.4pt,
                  minimum size=4.8pt,inner sep=0pt},
  tgtnode/.style={anchor=west,rounded corners=2.5pt,fill=ctgt!8,draw=ctgt,
                  line width=0.6pt,text=ctgtlabeltext,inner xsep=3.5pt,
                  inner ysep=2.2pt,font=\ttfamily\fontsize{6.2}{7}\selectfont\bfseries},
  depedge/.style={draw=cdepedge,line width=0.4pt,opacity=0.45,line cap=round},
  tgtedge/.style={draw=ctgtedge,line width=0.7pt,opacity=0.6,line cap=round},
  tfine/.style={fill opacity=0.28,draw opacity=0.36,minimum size=3pt,line width=0.2pt},  
}

\def\PFcycle{26}
\def\peakunclosed{6136}

\def\nlwfinal{15196}
\def\totalfinal{156}

\def\ndef{36}
\def\nthm{119}
\def\ntarget{5}
\def\nclosure{4}
\def\ncoarse{47}
\def\coarsetotal{20}
\def\coarsepfstart{26}
\def\coarseplatstart{80}
\def\coarseplatend{101}

\hypersetup{colorlinks=true,linkcolor=blue!50!black,citecolor=blue!50!black,urlcolor=blue!50!black}
\setlist[itemize]{leftmargin=1.25em,itemsep=1pt,topsep=2pt}
\setlist[enumerate]{leftmargin=1.35em,itemsep=1pt,topsep=2pt}
\usepackage{floatpag}\floatpagestyle{empty} 
\usepackage[normalem]{ulem} 

\newcommand{\word}[1]{\textbf{\emph{#1}}}
\newcommand{\system}{Trellis\xspace}

\newcommand{\lean}{Lean}
\newcommand{\tla}{TLA$^{+}$}

\newcommand{\tightparagraph}[1]{\vspace{2pt}\noindent\textbf{#1}}
\usepackage{dblfloatfix} 

\begin{document}

\begin{center}
\hrule height 1pt\vspace{0.23in}
{\Large\bfseries (Auto)formalization is supposed to be easy:\\
\system{} process semantics for spelling out rigorous proofs}\vspace{0.22in}

{\large Wesley Pegden\footnote{Department of Mathematical Sciences, Carnegie Mellon University.}}\vspace{0.2in}

\hrule height 1pt\vspace{0.28in}
\end{center}

\begin{abstract}
We present \system{}: an autoformalization system that leverages LLM agents in a deterministically constrained workflow to enforce incremental progress in Lean autoformalization tasks through iterative refinement of natural language proofs. Our approach is motivated by the common mathematician's notion of what it means to have a rigorous proof in the first place: namely, that it would be routine to elaborate any part of the proof in further detail. The result is a system which aims to achieve reliable autoformalization on a modest budget and with generalist agents, with specialization to autoformalization coming not from any task-specific agent training but instead from a meaning-of-rigor inspired workflow enforced by process semantics. We link to an end-to-end \lean{} formalization of a recent Ramsey theory breakthrough produced by the process.
\end{abstract}

\section{Introduction}
Formalism underpins notions of rigor in mathematics, but usually not directly; when mathematicians talk of having a rigorous proof of a theorem, it is quite rarely because they have written out a specific type of proof object in a particular formal system. Instead, when claiming an argument is mathematically rigorous, the claim is really instead that it is possible to ``spell out all the details'' of the argument to any required level. Indeed, the implicit claim is not even that it would just be possible to elaborate a given rigorous proof into a fully formal one, but that it should in some sense be routine to do so.

  One promise of autoformalization is that the tedious work admitted by this claim can be alleviated by machines---that we should really be able to translate any rigorous mathematical argument into a truly formal one.  On the other hand, however, the very premise that formalization of a rigorous argument should be ``routine but tedious'' should imply that autoformalization itself should be an unremarkable, completely tractable task not just when machines are brilliant but once they can follow and understand a correct argument.  That is, autoformalization should be routine \emph{today}, with existing publicly available LLM agents, on a modest budget, on problems of real substance.

  We present \system{}, a process semantics for autoformalization motivated by this view of why formalization should be routine (rather than motivated by, say, the observation that LLM agents are generally good at coding). In particular, \system{} works with an off-the-shelf generalist LLM, but in a highly structured process intended to mirror the methodical proof-decomposition workflow a mathematician might imagine when claiming that indeed, all the details can be filled in.  The result is a workflow in which agents build and then maintain and expand an end-to-end proof structure held together by natural language proofs, as the process formalizes or refines individual pieces of this structure step-by-step.

  The fundamental problem this is an answer to is how to enforce \emph{incremental progress} during an autoformalization task. Our answer is that progress can be enforced in our framework for the same reason that (auto)formalization is supposed to be easy: we can always refine parts of our proof until they are so spelled out that formalizing that piece is an easy task. The \system{} framework is designed to enforce a notion of refinement that maintains faithful correspondence to the formalization blueprint: the rigorous proof being formalized.

  \tightparagraph{Two complete Lean formalizations} of recent Ramsey theory breakthroughs are now publicly available alongside this draft, at \url{https://math.cmu.edu/~wes/trellis.php}: \emph{Improving $R(3,k)$ in just two bites} by Zion Hefty, Paul Horn, Dylan King, Florian Pfender~\cite{twobites}, and \emph{Nearly tight exponents for off-diagonal Ramsey numbers} by Domagoj Brada\v{c}~\cite{bradac}.  The former achieves a conjectured-asymptotically-optimal $R(3,k)$ lower bound; the latter breaks a 50-year-old exponent barrier for the general off-diagonal case.  At the linked site, we provide viewers which allow easy browsing of both the Lean code generated by Trellis, and the natural language ``glue'' that holds the Trellis proof structure together.

  The formalization of the \emph{two bites} paper took place over a period of two weeks when Trellis was under heavy development and thus when features and behavior were shifting from day to day.  At the end of the period, the autoformalization had reached a nearly complete state, but with a structure incompatible with newer Trellis features, so it was finished manually at the end: a human operator directed Codex to complete the repair along the paper-faithful route the system had already identified, without supplying mathematical content, Lean proof steps, or formalization-specific hints. The public formalization repo includes checkpoints before and after this manual edit.

  The formalization of the Brada\v{c} paper took place with a much more settled codebase and thus we release the full end-to-end autoformalization repo~\cite{offdiagonal_formalized} in which every cycle-checkpoint of the process is visible---that is, it is possible to inspect every edit made by the Trellis worker in the formalization process.  This autoformalization finished in roughly two days, using 35\% of the weekly usage budget of a ChatGPT Pro subscription; it completed on May~30, 2026, three days after Brada\v{c}'s paper first appeared on the arXiv.

%

\section{Failure modes motivating the \system{} design}
\label{sec:failure-modes}

In early experiments with a range of simpler autoformalization schemes, we observed a range of failure modes; i.e., ways in which the process would claim to formalize an intended paper target when that claim was actually not justified.

The simplest of these failure modes is simply a process claiming to have a complete \lean{}~\cite{lean} proof of something when it doesn't; e.g., perhaps \lean{} does successfully build the project, but `sorry' placeholders are used in various places in place of complete proofs.  This is a failure mode that can easily arise in the simplest of experiments, where a deterministic script simply repeatedly asks an agent to continue until it reports that it is done (in which case the process terminates); an imperfect agent may well report done before it is truly finished. Devious instances of this problem class could involve using unauthorized axioms or tampering with the build environment. The natural solution to this problem is to have the autoformalization process independently check all \lean{} build claims, in an environment agents don't have access to.

Another common problem facing autoformalization schemes is the possibility that agents in an autoformalization process do formalize \emph{something}, but not something corresponding in meaning to the actual formalization target. In particular, this problem can easily arise even with a strong adversarial model of \lean{} checking.  In one target in an early experiment, the quantity of interest was the \emph{maximum} size of a family.  Early experiments even with strong agents frequently revealed cases where one unfaithful buried definition undermined the meaning of all paper targets.  The natural solution to this problem is to have a system fix a set of approved target \lean{} statements before autoformalization begins.


After solving the first two problems, the risk is no longer false or unfaithful formalizations.  The remaining problem one sees with weak systems is a \textbf{failure to enforce real progress}.  One example of work without progress in our experiments running hardened but simplistic systems on hard problems is empty wrapping.  An agent makes ``progress'' by telling their reviewing agent, ``I've reduced the problem to this,'' ``I've reduced the problem to that,'' etc., indefinitely deferring some essential part of the mathematical work to lemmas with increasingly long names. In one experiment, a single target proof file grew to thousands of lines, in which essentially every lemma derived its conclusion from an unproved hypothesis, and successive cycles renamed and repackaged that hypothesis---without ever
reducing the open-assumption set---over roughly one hundred and fifty cycles.

In \system{}, we address the first problem by adopting an adversarial stance for checking \lean{} builds; all \lean{} builds are carried out by the deterministic supervisor in an isolated environment the editing agent doesn't have access to.  We address the second problem by having the process settle \lean{} statements (including relevant definitions) relevant to paper targets before formalization; these settled statements are offered for human review and no changes to them are allowed during formalization without reopening human review.

The third problem is the most interesting of the three problems; simple metrics like line counts and sorry counts are poor guides. Unproductive wrapping is great at increasing lines of code, and, of course, it only takes one sorry to take the place of the proof of any paper target. It is our solution to \emph{this} problem that is motivated by our meaning-of-rigor viewpoint, that for rigorous natural language proofs, it should always be feasible to elaborate them to any desired level of detail. We enforce progress by enforcing that the process is always increasing meaningful detail in a proof's decomposition.

\section{Rigor as Latent Formality}
\label{sec:rigor}

The \system{} semantics rest on a working definition of what it means for a mathematical proof to be rigorous:
\begin{quote}
A rigorous proof is one for which it is routine to elaborate any step of the proof to any level of detail.
\end{quote}

The \system{} process is designed to operationalize this definition: to orchestrate imperfect agents acting under the rules of a deterministic supervisor in such a way that details really do get ``spelled out'', to the point that any part of the proof becomes easy to formalize.  Importantly, we leverage the ability of agents to work with mathematics in natural language as much (if not more) than we leverage their ability to write \lean{} code; mathematics in natural language is the glue holding together our tablet of nodes as our process elaborates the paper we are formalizing.

\system{} operates over a \emph{proof tablet}: a directed acyclic graph (defined by \lean{} import structure). A node is either a \emph{definition} or a \emph{theorem-like} statement (theorem, lemma, corollary, or helper), and every node carries two paired sides---a natural-language side written in \LaTeX{} and a \lean{} side.

To operationalize the notion that rigorous natural language proofs can always be ``elaborated'', \system{} builds and then maintains the invariant that the Tablet is in a state where every node has been judged by (separate) agents to satisfy the following three verification gates:
\label{sec:gates}
\begin{enumerate}[label=(\alph*)]
\item \textbf{Substantiveness:} (Natural Language) The \LaTeX{} side of the node: (1) genuinely matches a statement used (explicitly or implicitly) by the proof described by the paper, and (2) represents a meaningful refinement of the current Tablet; that is, it is not essentially the same in meaning as any other current node.
\item \textbf{Correspondence:} (Natural Language and Lean): The \lean{} side of the node genuinely corresponds to the mathematical meaning of the \LaTeX{} side of the node.
\item \textbf{Soundness:} (Natural Language): The \LaTeX{} side of the proof contains a line-by-line checkable natural-language proof with all needed dependencies available in Mathlib or in Tablet nodes imported by the node.
\end{enumerate}

The first phase of the \system{} process is a \emph{Theorem-stating phase} in which an initial directed acyclic graph (DAG) of \system{} nodes is built and refined until all nodes satisfy these three verifications.  A fourth verifier lane \textbf{Paper-faithfulness} verifies that nodes that purport to cover target theorems in the paper being formalized truly do so.  But we do not want to trust agent quality for inferring correctness of formalization, so \system{} computes the list of all nodes whose Lean statements affect the meaning of nodes that purport to cover paper targets, and presents these nodes for human acceptance.  Only after this does the process proceed to the formalization phase\footnote{No changes are allowed to the human-approved semantic-relevant nodes without reopening human approval.}.

In the proof formalization phase, an ideal step consists of writing a \lean{} proof for the \lean{} statement of a node, from its imports.  \system{} then evaluates (in a separate environment the worker doesn't have access to) whether this proof is valid conditioned on its inputs and doesn't use any unapproved axioms.  A fundamental reason autoformalization is hard, however, is that an agent will not always succeed at writing such a proof; frequently, this task will be considered too difficult to do all at once.  In this situation, the \system{} process allows the agent another option: you may meaningfully enrich the DAG below this node, e.g., by adding helper nodes, \emph{but they will be required to pass the verification lanes \textbf{substantiveness}, \textbf{correspondence}, and \textbf{soundness}} (soundness is waived for nodes already \lean{}-closed). The promise of this rule is that while agents may sometimes make mistakes, they should not systematically work in an unproductive way.  When a worker refines the proof DAG rather than closing a \lean{} proof, it must, in essence, produce natural language certificates (checked by independent agents) that demonstrate its refined decomposition is meaningful (and, via \textbf{substantiveness} of new nodes, truly represents a refinement, rather than just the addition of pointless wrapper nodes).  If a worker can only meaningfully refine a proof-step or formalize it, the \system{} philosophy goes, the worker cannot help but get to a point where formalization of one step of the proof eventually becomes a manageable task.

\subsection{Contract enforcement rather than prompt engineering}
\label{sec:determinism}

One could imagine trying to build an autoformalization system with the same motivation we describe here, engineered entirely out of a rotating sequence of prompts (e.g., in simplest form: to Agent 1: work on the Tablet: to Agents 2,3,4,\dots: Verify the work on the Tablet respects the verification lane rules, to Agent 1: Read the feedback from those agents and continue to work on the Tablet). In our experience, off-the-shelf agents still do not currently have the long-term horizon/perspective for this workflow to be reliable on difficult tasks orders of magnitude longer than the context window of any one agent. 

In particular, a workflow built only from agent prompts would let the judgments of verifier lanes from the distant past (but still for live Tablet nodes) be quietly bypassed.  In \system{}, a reviewer agent directs the actions of a worker agent, but they are \emph{both} tightly constrained by a deterministic kernel (leveraging durable long-term state memory) that owns semantic authority of the process.  It tracks not only Lean build status but the results of verifier lanes, routing scope, Tablet permissions and state, etc.  The \system{} reviewer will never tell the \system{} worker to work on the \lean{} proof of a node that has not passed correspondence, because the deterministic kernel will reject such a directive and tell the reviewer to produce another one.  And if a \system{} worker edits the \lean{} statement of a node to address a correspondence issue, it is the kernel that computes via \lean{} the set of upstream nodes whose own correspondence must be reopened for verification because of this change. The tablet is not merely a file layout: it is the progress ledger, and a local edit counts as progress only when it advances the tablet in a way allowed by the kernel-authorized scope chosen by the reviewer agent.

That the two sides of a node are concrete files with a fixed shape is what makes their relationship checkable rather than conventional. The natural-language side of a node is a single statement block followed (except in the case of definition nodes) by a single proof block; the \lean{} side is a single principal declaration whose name matches the node, with an explicit marker separating the statement from the proof body. Dependencies are cited by name in both \lean{} and \LaTeX{}, via imports in \lean{}, and \verb|\noderef{}| commands in \LaTeX{}. Deterministic structural checks enforce this shape, and the shape is then used by the deterministic kernel to decide when, e.g., soundness verification has to be reopened for nodes because of downstream changes to statements.

\subsection{Actors}
\system{} separates roles that agentic workflows often conflate.
\emph{Workers} propose statements, dependencies, decompositions, and proof edits, and can also request different scope in the face of problems.  \emph{Verifier lanes} evaluate the semantic relations of \S\ref{sec:gates}; each lane is independent of the worker that produced the change and is judged by a single agent.\footnote{A lane may instead be run as a multi-vote \emph{panel}, where the kernel only records unanimous decisions and leaves split panels as \word{Unknown}.} \emph{The reviewer} is a single adjudicator that consumes worker outcomes, accumulated verifier evidence, and the open-blocker set, and decides whether to continue locally, restructure, assign a blocker as a task, or escalate, \emph{within the confines of the rules enforced by the deterministic kernel}. \emph{The human operator} has a narrow task: provide the paper being formalized, and approve the \lean{} shape of the kernel-computed set of nodes relevant to paper targets at the phase transition from theorem stating to proof formalization.

\section{Recursive Refinement in Practice}
\label{sec:practice}

\begin{figure}[t]
\small
\setlength{\abovedisplayskip}{2pt}\setlength{\belowdisplayskip}{2pt}%
\setlength{\abovedisplayshortskip}{1pt}\setlength{\belowdisplayshortskip}{1pt}
\noindent\hrulefill\\[1pt]
\noindent{\ttfamily\bfseries MainTheorem}\hfill{\footnotesize\itshape paper target}\\[1pt]
For any \(s\ge 4\) there is a constant \(c_s>0\) such that for all \(k\ge 2\),
\[ r(s,k)\ \ge\ c_s\,\frac{k^{\,s-2}}{(\log k)^{\,2s-6}}. \]
\begin{lstlisting}[style=lean]
theorem MainTheorem :
  ∀ s : ℕ, 4 ≤ s → ∃ c : ℝ, 0 < c ∧
    ∀ k : ℕ, 2 ≤ k →
      c * ((k:ℝ)^(s-2)) / ((Real.log (k:ℝ))^(2*s-6))
        ≤ (RamseyNumber s k : ℝ)
\end{lstlisting}
\vspace{1pt}\noindent\hrulefill\\[1pt]
\noindent{\ttfamily\bfseries RamseyNumber}\hfill{\footnotesize\itshape semantic-closure definition}\\[1pt]
The two-color Ramsey number \(r(s,k)\) is the least \(n\) such that every graph on \(n\) vertices contains a clique of size \(s\) or an independent set of size \(k\).
\begin{lstlisting}[style=lean]
noncomputable def RamseyNumber (s k : ℕ) : ℕ :=
  sInf {n : ℕ | RamseyProperty s k n}
\end{lstlisting}
\vspace{1pt}\noindent\hrulefill\\[1pt]
\noindent{\ttfamily\bfseries RamseyProperty}\hfill{\footnotesize\itshape semantic-closure definition}\\[1pt]
\(R(s,k,n)\): every simple graph on \(n\) vertices contains a clique of size \(s\) or an independent set of size \(k\).
\begin{lstlisting}[style=lean]
def RamseyProperty (s k n : ℕ) : Prop :=
  ∀ G : SimpleGraph (Fin n),
    (∃ S : Finset (Fin n), G.IsNClique s S) ∨
      (∃ I : Finset (Fin n), Gᶜ.IsNClique k I)
\end{lstlisting}
\noindent\hrulefill
\caption{The human-protected core of one of the five targets (Figure~\ref{fig:tablet}) from Brada\v{c}'s paper: the main theorem's statement and \lean{} declaration, with the two definitions in its kernel-computed semantic closure---the only nodes the target's meaning depends on, and the only \lean{} shapes a human ratifies.  The faithfulness and correctness of the final formalization is only subject to the faithfulness of the nodes in this semantic closure, without any assumptions on the reliability of LLM agents.}
\label{fig:targetcore}
\end{figure}

We illustrate the process on a single end-to-end run: an unguided formalization of Brada\v{c}'s recent paper \emph{Nearly tight exponents for off-diagonal Ramsey numbers}~\cite{bradac}. The paper is short but cutting-edge, improving the half-century-old Spencer lower bound on off-diagonal Ramsey numbers, showing $R(s,k)\ge c_s\, k^{s-2}/(\log k)^{2s-6}$ for $s\ge 4$, via an algebraic construction (polarity graphs of projective spaces) whose independent sets are counted with the container method. We pointed \system{} at five distinct headline theorems of the paper (\texttt{thm:main}, \texttt{thm:off-diagonal-general}, \texttt{thm:k-Ck}, \texttt{thm:close}, and the multicolor bound \texttt{thm:multicolor}) and otherwise left it alone. A single off-the-shelf general-purpose reasoning model filled every role---worker, reviewer, and all four verifier lanes---with no task-specific training.\footnote{The model used here was GPT-5.5 xhigh, which likewise filled all roles in the other formalization linked with this manuscript~\cite{twobites_formalized}. \system{} has also been run successfully with Opus 4.7 max and Gemini 3.1-pro preview in various roles.} This run never invoked the separate per-file deviation lane of \S\ref{sec:deviations}: it followed the paper's exact path throughout, needing no authorized departure.

\tightparagraph{Outcome.} The run halted cleanly after $140$ supervisor cycles with all five paper targets formalized. The terminal tablet has $156$ nodes ($119$ theorem-like, $36$ definitions, one preamble) totaling $12{,}854$ lines of \lean{}. Every one of the $119$ proof obligations builds in the supervisor's isolated checker with no \verb|sorryAx| or other nonstandard axiom dependencies.

\tightparagraph{Refinement, not redefinition.} Figure~\ref{fig:progress} tracks three quantities across the whole run: the total node count, the number of nodes passing every verifier lane (correspondence, substantiveness, and soundness), and the number of nodes \lean{}-closed in the isolated checker. Two regimes are visible. In theorem-stating (through cycle~$\sim$25) the tablet settles at $48$ nodes and the verifier-lane curve climbs to meet the total, as drafted statements pass their gates; the human then approves the target-semantic \lean{} shapes and the run crosses into formalization. There, the central observation is the shape of the \lean{}-closure curve. Our process semantics prohibit the lane curve from straying indefinitely from the node total, as every introduced node is required to clear substantiveness, correspondence, and soundness as it is added. In particular, when a worker cannot close a \lean{} obligation outright, its only sanctioned alternative is to spell the step out further, and substantiveness (Clause~2) forbids the spelling-out from being a vacuous repackaging. The growth is genuine decomposition, not the empty wrapping that defeated the simpler pipelines of \S\ref{sec:failure-modes}. Table~\ref{tab:lane-stats} quantifies the pass rate of verifier lanes in the Brada\v{c} paper run. 

\begin{figure}[t]
\centering
\resizebox{\linewidth}{!}{%
\begin{tikzpicture}
\begin{axis}[
    width=\linewidth, height=5.7cm,
    xmin=1, xmax=143, ymin=0, ymax=170,
    xtick={1,20,40,60,80,100,120,140},
    xlabel={supervisor cycle}, ylabel={nodes},
    axis lines=left, tick align=outside,
    every axis plot/.append style={no markers},
    xlabel near ticks, ylabel near ticks,
    label style={font=\small}, tick label style={font=\footnotesize},
    clip mode=individual,
    legend style={at={(0.02,0.98)}, anchor=north west, draw=none, fill=none,
                  font=\fontsize{6.6}{7.6}\selectfont, row sep=-1pt,
                  legend cell align=left},
]
  \addplot[draw=none, fill=cthm, forget plot] table[x=cycle,y=total] {figs/progress.dat} \closedcycle;
  \addplot[draw=none, fill=cdef, forget plot] table[x=cycle,y=defs]  {figs/progress.dat} \closedcycle;
  \addplot[ctotal, line width=0.9pt, forget plot] table[x=cycle,y=total] {figs/progress.dat};
  \addplot[clane,  line width=1.2pt, forget plot] table[x=cycle,y=lanes] {figs/progress.dat};
  \addplot[clean,  line width=1.6pt, forget plot] table[x=cycle,y=lean]  {figs/progress.dat};
  \draw[cphase, dashed, line width=0.7pt] (axis cs:\PFcycle,0) -- (axis cs:\PFcycle,170);
  \node[cphase, font=\fontsize{6}{6}\selectfont, anchor=south east, align=right]
        at (axis cs:\PFcycle,4) {thm-\\stating};
  \node[cphase, font=\fontsize{6}{6}\selectfont, anchor=south west]
        at (axis cs:\PFcycle,4) {proof formalization};
  \node[ctotal, font=\fontsize{6.8}{7}\selectfont, anchor=west] at (axis cs:141,156) {\totalfinal};
  \addlegendimage{ctotal, line width=1.4pt}        \addlegendentry{total nodes}
  \addlegendimage{clane, line width=1.4pt}         \addlegendentry{passing all verifier lanes}
  \addlegendimage{clean, line width=1.8pt}         \addlegendentry{Lean-closed}
  \addlegendimage{area legend, draw=none, fill=cthm} \addlegendentry{theorem-like}
  \addlegendimage{area legend, draw=none, fill=cdef} \addlegendentry{definitions}
\end{axis}
\end{tikzpicture}}
\caption{Verification and \lean{}-closure progress over the run. Total nodes (definitions in blue, theorem-like in orange) and the count passing \emph{all} verifier lanes (solid) rise together---every committed node is lane-clean---while \lean{}-closure (black) lags, catching the total only at termination. The human-approved scaffold barely moves ($28\to37$); the theorem-like population grows $20\to119$.}
\label{fig:progress}

\vspace{10pt}

\resizebox{\linewidth}{!}{%
\begin{tikzpicture}
\begin{axis}[
    width=\linewidth, height=5.7cm,
    xmin=1, xmax=143, ymin=0, ymax=18600,
    xtick={1,20,40,60,80,100,120,140},
    xlabel={supervisor cycle}, ylabel={natural-language proof words},
    axis lines=left, tick align=outside,
    every axis plot/.append style={no markers},
    xlabel near ticks, ylabel near ticks,
    label style={font=\small}, tick label style={font=\footnotesize},
    scaled y ticks=false,
    ytick={0,5000,10000,15000},
    yticklabel style={/pgf/number format/fixed, /pgf/number format/1000 sep={,}},
    extra y ticks={\peakunclosed},
    extra y tick style={tick label style={text=cred, font=\footnotesize}},
    clip mode=individual,
    legend style={at={(0.02,0.98)}, anchor=north west, draw=none, fill=none,
                  font=\fontsize{6.6}{7.6}\selectfont, row sep=-1pt,
                  legend cell align=left},
]
  \addplot[draw=none, fill=cgreenfill, forget plot] table[x=cycle,y=nlw]   {figs/words.dat} \closedcycle;
  \addplot[draw=none, fill=credfill,   forget plot] table[x=cycle,y=nlunc] {figs/words.dat} \closedcycle;
  \addplot[cgreen, line width=1.3pt, forget plot] table[x=cycle,y=nlw]   {figs/words.dat};
  \addplot[cred,   line width=1.0pt, forget plot] table[x=cycle,y=nlunc] {figs/words.dat};
  \draw[cphase, dashed, line width=0.7pt] (axis cs:\PFcycle,0) -- (axis cs:\PFcycle,18600);
  \node[cphase, font=\fontsize{6}{6}\selectfont, anchor=south east, align=right]
        at (axis cs:\PFcycle,250) {thm-\\stating};
  \node[cphase, font=\fontsize{6}{6}\selectfont, anchor=south west]
        at (axis cs:\PFcycle,250) {proof formalization};
  \node[cgreen, font=\fontsize{6.8}{7}\selectfont, anchor=west] at (axis cs:141,\nlwfinal) {15{,}196};
  \addlegendimage{cgreen, line width=1.5pt} \addlegendentry{total NL proof words}
  \addlegendimage{area legend, draw=none, fill=credfill}   \addlegendentry{words in un-Lean-closed nodes}
  \addlegendimage{area legend, draw=none, fill=cgreenfill} \addlegendentry{words backed by Lean}
\end{axis}
\end{tikzpicture}}
\caption{The natural-language proof corpus through the elaboration process. By the end of theorem stating the total \LaTeX{} proof words (green) already exceed $6{,}000$---more than the target paper's ${\approx}3{,}700$ (excluding introduction and bibliography). Through formalization the total grows to $15{,}196$, while the share in \emph{un-\lean{}-closed} nodes (red) trends to zero: every proof obligation ends supported by a valid \lean{} proof. The red tick marks the peak un-\lean{}-closed backlog of $6{,}136$ words.}
\label{fig:words}
\end{figure}

\begin{table}[t]
\centering
\footnotesize
\setlength{\tabcolsep}{4pt}
\renewcommand{\arraystretch}{1.1}
\begin{tabular}{@{}l rr rr r@{}}
\toprule
& \multicolumn{2}{c}{\textit{theorem stating}} & \multicolumn{2}{c}{\textit{proof formalization}} & \\
\cmidrule(lr){2-3}\cmidrule(lr){4-5}
\textbf{Lane} & calls & pass & calls & pass & pass-rate \\
\midrule
substantiveness & $17$ & $14$ & $117$ & $113$ & $94.8\%$ \\
correspondence  & $20$ & $15$ & $113$ & $109$ & $93.2\%$ \\
soundness       & $21$ & $19$ & $4$   & $4$   & $92.0\%$ \\
\midrule
total           & $58$ & $48$ & $234$ & $226$ & $93.8\%$ \\
\bottomrule
\end{tabular}
\caption{Per-node verifier-lane activity over the run on the Brada\v{c} paper, split by
phase. A \emph{call} is one verdict the lane returned for a node; \emph{pass}
counts the \texttt{Pass} verdicts among them (re-verification after a
fingerprint drift is a fresh call). Note that soundness verification is waived for nodes introduced already lean-closed, as was the case with the majority of new nodes introduced in this run during the proof-formalization phase.}
\label{tab:lane-stats}
\end{table}%

\tightparagraph{From prose to \lean{}.} Figure~\ref{fig:words} views the same run through its natural-language proofs. \system{} treats \LaTeX{} proofs as the load-bearing glue---the soundness lane demands a line-by-line checkable argument for every node. The total grows to $15{,}196$ words as statements are decomposed and justified; for comparison Brada\v{c}'s paper is approximately $3{,}700$ words, excluding the introduction and bibliography. The diagnostic curve is the \emph{un-\lean{}-closed} share, which can be taken as a crude measure of the unfinished \lean{} obligation; note that it is a unique feature of our method based on natural-language elaboration that we get this kind of concrete metric of remaining work.

The decomposition is visibly recursive. The main theorem, for instance, was not closed in one shot; the worker reduced it through a chain of substantive helpers it had to state and justify along the way---a dyadic Galois-field scaling lemma, the polarity-graph setup and parameter bounds, an eigenvalue bound, a large-$k$ comparison, and a finite-range absorption step---each a real sub-obligation that itself cleared the gates. The technical heart of the argument, counting independent sets in the construction, accreted the largest single cluster of machinery (a family of some two dozen ``bad-tuple'' counting nodes). The last recorded worker action is representative: it closed \texttt{thm:off-diagonal-general} by introducing one final \lean{}-closed numerical-absorption helper and using it for the range and Ramsey comparisons. 

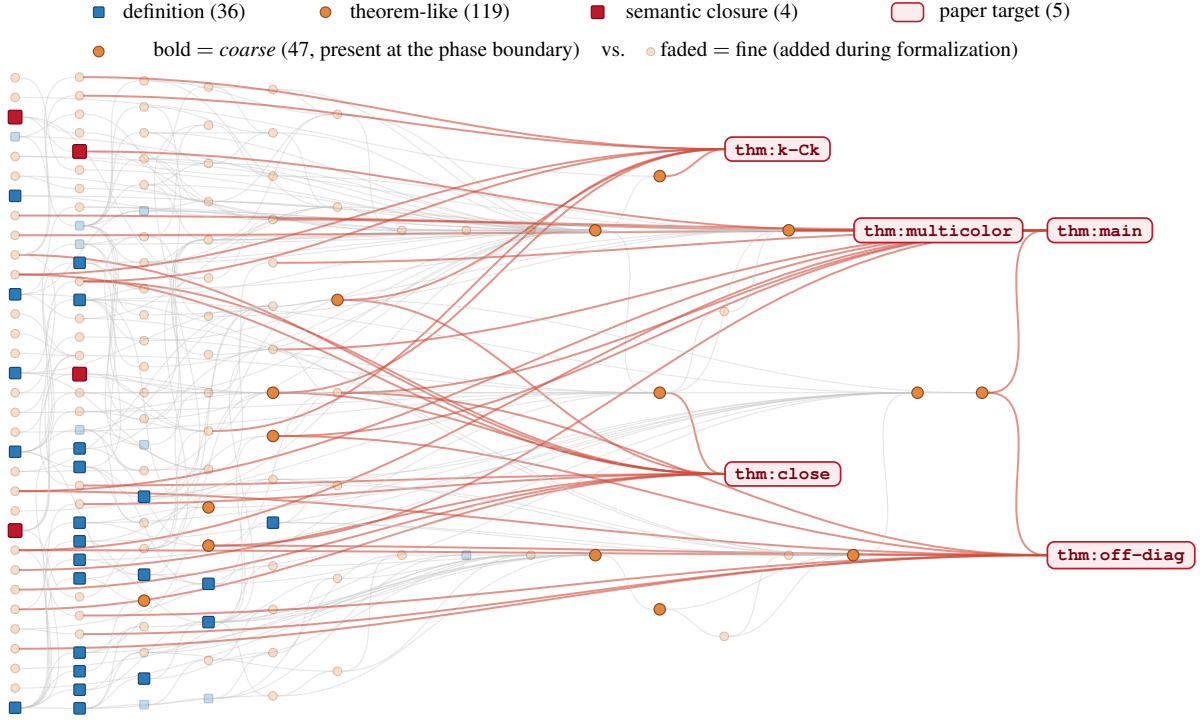
\begin{figure*}[tbp]
\centering
\begin{tikzpicture}[baseline]
  \node[defnode] at (0,0) {};
  \node[anchor=west, font=\footnotesize] at (0.2,0) {definition (\ndef)};
  \node[thmnode] at (3.0,0) {};
  \node[anchor=west, font=\footnotesize] at (3.2,0) {theorem-like (\nthm)};
  \node[closurenode] at (6.6,0) {};
  \node[anchor=west, font=\footnotesize] at (6.85,0) {semantic closure (\nclosure)};
  \node[rounded corners=2pt, fill=ctgt!8, draw=ctgt, line width=0.6pt,
        minimum width=12pt, minimum height=7pt, inner sep=0pt] at (10.7,0) {};
  \node[anchor=west, font=\footnotesize] at (11.0,0) {paper target (\ntarget)};
  \node[thmnode] at (0,-0.52) {};
  \node[anchor=west, font=\footnotesize] at (0.6,-0.52)
       {bold $=$ \emph{coarse} (\ncoarse, present at the phase boundary)\quad vs.\quad \tikz[baseline=-0.5ex]{\node[thmnode,tfine]{};}~faded $=$ fine (added during formalization)};
\end{tikzpicture}

\resizebox{0.99\textwidth}{!}{%
\begin{tikzpicture}[x=1cm,y=1cm]
  \node[defnode] (n0) at (0.350,5.612) {};
  \node[thmnode,tfine] (n1) at (1.144,6.686) {};
  \node[thmnode,tfine] (n2) at (1.144,5.314) {};
  \node[thmnode,tfine] (n3) at (0.350,1.733) {};
  \node[thmnode,tfine] (n4) at (0.350,1.976) {};
  \node[thmnode,tfine] (n5) at (1.144,3.029) {};
  \node[thmnode,tfine] (n6) at (0.350,5.855) {};
  \node[thmnode,tfine] (n7) at (0.350,2.218) {};
  \node[tgtnode] (n8) at (9.081,3.400) {thm:close};
  \node[thmnode,tfine] (n9) at (0.350,6.097) {};
  \node[thmnode,tfine] (n10) at (1.144,3.257) {};
  \node[thmnode] (n11) at (2.731,2.518) {};
  \node[defnode] (n12) at (0.350,3.673) {};
  \node[defnode] (n13) at (1.144,6.000) {};
  \node[defnode,tfine] (n14) at (1.144,3.943) {};
  \node[thmnode,tfine] (n15) at (1.938,3.440) {};
  \node[thmnode,tfine] (n16) at (3.525,5.467) {};
  \node[thmnode,tfine] (n17) at (2.731,3.459) {};
  \node[thmnode,tfine] (n18) at (1.938,2.800) {};
  \node[thmnode] (n19) at (4.319,5.543) {};
  \node[thmnode] (n20) at (7.494,2.400) {};
  \node[defnode,tfine] (n21) at (0.350,7.552) {};
  \node[thmnode,tfine] (n22) at (2.731,4.871) {};
  \node[thmnode,tfine] (n23) at (6.700,6.400) {};
  \node[thmnode,tfine] (n24) at (5.906,6.400) {};
  \node[defnode,tfine] (n25) at (1.938,6.640) {};
  \node[thmnode,tfine] (n26) at (5.112,6.400) {};
  \node[thmnode,tfine] (n27) at (0.350,7.067) {};
  \node[thmnode,tfine] (n28) at (4.319,6.686) {};
  \node[thmnode,tfine] (n29) at (2.731,6.753) {};
  \node[thmnode,tfine] (n30) at (1.144,6.914) {};
  \node[thmnode,tfine] (n31) at (4.319,7.829) {};
  \node[thmnode,tfine] (n32) at (2.731,8.165) {};
  \node[thmnode,tfine] (n33) at (3.525,8.133) {};
  \node[thmnode,tfine] (n34) at (3.525,7.600) {};
  \node[thmnode,tfine] (n35) at (2.731,7.224) {};
  \node[thmnode,tfine] (n36) at (1.938,7.280) {};
  \node[thmnode,tfine] (n37) at (1.144,7.829) {};
  \node[thmnode,tfine] (n38) at (1.938,7.600) {};
  \node[defnode,tfine] (n39) at (1.144,6.457) {};
  \node[thmnode,tfine] (n40) at (1.938,4.080) {};
  \node[defnode,tfine] (n41) at (1.938,3.760) {};
  \node[thmnode,tfine] (n42) at (2.731,6.282) {};
  \node[thmnode,tfine] (n43) at (1.938,4.400) {};
  \node[thmnode,tfine] (n44) at (1.938,7.920) {};
  \node[thmnode,tfine] (n45) at (1.938,8.240) {};
  \node[thmnode,tfine] (n46) at (1.938,6.960) {};
  \node[thmnode,tfine] (n47) at (1.938,6.000) {};
  \node[thmnode,tfine] (n48) at (1.938,4.720) {};
  \node[thmnode,tfine] (n49) at (1.938,5.040) {};
  \node[thmnode,tfine] (n50) at (0.350,8.279) {};
  \node[thmnode,tfine] (n51) at (1.144,7.143) {};
  \node[thmnode] (n52) at (8.287,7.067) {};
  \node[thmnode] (n53) at (8.287,4.400) {};
  \node[thmnode,tfine] (n54) at (0.350,8.036) {};
  \node[thmnode] (n55) at (7.494,6.400) {};
  \node[thmnode,tfine] (n56) at (0.350,4.885) {};
  \node[thmnode,tfine] (n57) at (0.350,3.915) {};
  \node[thmnode,tfine] (n58) at (0.350,4.400) {};
  \node[thmnode,tfine] (n59) at (0.350,1.491) {};
  \node[thmnode,tfine] (n60) at (1.144,4.857) {};
  \node[thmnode,tfine] (n61) at (0.350,5.127) {};
  \node[thmnode,tfine] (n62) at (0.350,4.158) {};
  \node[thmnode,tfine] (n63) at (0.350,5.370) {};
  \node[thmnode,tfine] (n64) at (1.144,5.086) {};
  \node[thmnode,tfine] (n65) at (0.350,2.945) {};
  \node[thmnode,tfine] (n66) at (0.350,0.764) {};
  \node[thmnode,tfine] (n67) at (0.350,1.006) {};
  \node[defnode] (n68) at (1.144,3.486) {};
  \node[defnode] (n69) at (1.938,3.120) {};
  \node[defnode] (n70) at (1.144,2.800) {};
  \node[thmnode,tfine] (n71) at (1.144,8.057) {};
  \node[tgtnode] (n72) at (9.081,7.400) {thm:k-Ck};
  \node[thmnode,tfine] (n73) at (1.144,8.286) {};
  \node[defnode] (n74) at (0.350,0.521) {};
  \node[defnode] (n75) at (1.144,0.514) {};
  \node[thmnode,tfine] (n76) at (1.938,1.200) {};
  \node[thmnode,tfine] (n77) at (2.731,1.106) {};
  \node[defnode,tfine] (n78) at (1.938,0.560) {};
  \node[thmnode,tfine] (n79) at (5.112,2.400) {};
  \node[thmnode,tfine] (n80) at (3.525,1.733) {};
  \node[defnode] (n81) at (1.144,3.714) {};
  \node[thmnode,tfine] (n82) at (3.525,4.933) {};
  \node[defnode] (n83) at (1.144,2.343) {};
  \node[defnode] (n84) at (1.144,1.200) {};
  \node[thmnode,tfine] (n85) at (1.938,1.520) {};
  \node[defnode,tfine] (n86) at (2.731,0.635) {};
  \node[defnode] (n87) at (2.731,1.576) {};
  \node[thmnode,tfine] (n88) at (4.319,0.971) {};
  \node[thmnode,tfine] (n89) at (3.525,2.267) {};
  \node[thmnode,tfine] (n90) at (3.525,0.667) {};
  \node[defnode] (n91) at (1.938,0.880) {};
  \node[defnode] (n92) at (1.144,0.743) {};
  \node[thmnode,tfine] (n93) at (6.700,2.400) {};
  \node[defnode,tfine] (n94) at (5.906,2.400) {};
  \node[thmnode,tfine] (n95) at (3.525,1.200) {};
  \node[thmnode,tfine] (n96) at (4.319,2.114) {};
  \node[tgtnode] (n97) at (13.050,6.400) {thm:main};
  \node[thmnode,tfine] (n98) at (0.350,6.339) {};
  \node[thmnode,tfine] (n99) at (0.350,3.188) {};
  \node[thmnode,tfine] (n100) at (3.525,6.000) {};
  \node[thmnode,tfine] (n101) at (0.350,6.582) {};
  \node[thmnode,tfine] (n102) at (0.350,2.461) {};
  \node[thmnode,tfine] (n103) at (4.319,4.400) {};
  \node[closurenode] (n104) at (1.144,7.371) {};
  \node[thmnode,tfine] (n105) at (3.525,6.533) {};
  \node[closurenode] (n106) at (0.350,7.794) {};
  \node[thmnode,tfine] (n107) at (1.144,7.600) {};
  \node[thmnode,tfine] (n108) at (2.731,5.812) {};
  \node[tgtnode] (n109) at (10.669,6.400) {thm:multicolor};
  \node[defnode] (n110) at (0.350,6.824) {};
  \node[defnode] (n111) at (1.144,0.971) {};
  \node[thmnode,tfine] (n112) at (0.350,1.248) {};
  \node[thmnode,tfine] (n113) at (1.144,1.429) {};
  \node[thmnode,tfine] (n114) at (1.144,1.657) {};
  \node[tgtnode] (n115) at (13.050,2.400) {thm:off-diag};
  \node[defnode] (n116) at (1.144,2.114) {};
  \node[thmnode] (n117) at (3.525,3.867) {};
  \node[thmnode] (n118) at (1.938,1.840) {};
  \node[defnode] (n119) at (1.938,2.160) {};
  \node[defnode] (n120) at (3.525,2.800) {};
  \node[thmnode,tfine] (n121) at (4.319,3.257) {};
  \node[thmnode,tfine] (n122) at (9.875,2.400) {};
  \node[thmnode] (n123) at (11.462,4.400) {};
  \node[thmnode] (n124) at (10.669,2.400) {};
  \node[thmnode,tfine] (n125) at (9.081,1.400) {};
  \node[thmnode] (n126) at (2.731,2.988) {};
  \node[defnode] (n127) at (2.731,2.047) {};
  \node[defnode] (n128) at (1.144,2.571) {};
  \node[thmnode,tfine] (n129) at (3.525,3.333) {};
  \node[thmnode] (n130) at (12.256,4.400) {};
  \node[closurenode] (n131) at (1.144,4.629) {};
  \node[thmnode,tfine] (n132) at (2.731,3.929) {};
  \node[thmnode,tfine] (n133) at (2.731,4.400) {};
  \node[closurenode] (n134) at (0.350,2.703) {};
  \node[thmnode,tfine] (n135) at (1.144,4.171) {};
  \node[thmnode,tfine] (n136) at (1.144,4.400) {};
  \node[thmnode,tfine] (n137) at (1.938,2.480) {};
  \node[thmnode,tfine] (n138) at (1.144,1.886) {};
  \node[defnode,tfine] (n139) at (1.144,6.229) {};
  \node[thmnode] (n140) at (9.875,6.400) {};
  \node[thmnode,tfine] (n141) at (9.081,5.400) {};
  \node[thmnode,tfine] (n142) at (0.350,7.309) {};
  \node[thmnode,tfine] (n143) at (3.525,7.067) {};
  \node[thmnode,tfine] (n144) at (2.731,7.694) {};
  \node[thmnode,tfine] (n145) at (2.731,5.341) {};
  \node[thmnode,tfine] (n146) at (1.938,6.320) {};
  \node[thmnode,tfine] (n147) at (1.938,5.360) {};
  \node[thmnode,tfine] (n148) at (1.938,5.680) {};
  \node[thmnode] (n149) at (3.525,4.400) {};
  \node[defnode] (n150) at (0.350,4.642) {};
  \node[thmnode,tfine] (n151) at (1.144,5.771) {};
  \node[thmnode] (n152) at (8.287,1.733) {};
  \node[defnode] (n153) at (1.144,5.543) {};
  \node[thmnode,tfine] (n154) at (0.350,3.430) {};
  \begin{pgfonlayer}{bg}
    \draw[depedge] (n0) to[out=0,in=180] (n1);
    \draw[depedge] (n0) to[out=0,in=180] (n2);
    \draw[depedge] (n3) to[out=0,in=180] (n5);
    \draw[tgtedge] (n3) to[out=0,in=180] (n8);
    \draw[tgtedge] (n4) to[out=0,in=180] (n8);
    \draw[tgtedge] (n5) to[out=0,in=180] (n8);
    \draw[tgtedge] (n6) to[out=0,in=180] (n8);
    \draw[tgtedge] (n7) to[out=0,in=180] (n8);
    \draw[tgtedge] (n9) to[out=0,in=180] (n8);
    \draw[tgtedge] (n10) to[out=0,in=180] (n8);
    \draw[tgtedge] (n19) to[out=0,in=180] (n8);
    \draw[tgtedge] (n53) to[out=0,in=180] (n8);
    \draw[tgtedge] (n149) to[out=0,in=180] (n8);
    \draw[tgtedge] (n151) to[out=0,in=180] (n8);
    \draw[depedge] (n9) to[out=0,in=180] (n10);
    \draw[depedge] (n70) to[out=0,in=180] (n11);
    \draw[depedge] (n81) to[out=0,in=180] (n11);
    \draw[depedge] (n118) to[out=0,in=180] (n11);
    \draw[depedge] (n12) to[out=0,in=180] (n13);
    \draw[depedge] (n12) to[out=0,in=180] (n14);
    \draw[depedge] (n14) to[out=0,in=180] (n15);
    \draw[depedge] (n153) to[out=0,in=180] (n15);
    \draw[depedge] (n14) to[out=0,in=180] (n16);
    \draw[depedge] (n17) to[out=0,in=180] (n16);
    \draw[depedge] (n18) to[out=0,in=180] (n16);
    \draw[depedge] (n69) to[out=0,in=180] (n16);
    \draw[depedge] (n150) to[out=0,in=180] (n16);
    \draw[depedge] (n154) to[out=0,in=180] (n16);
    \draw[depedge] (n14) to[out=0,in=180] (n17);
    \draw[depedge] (n69) to[out=0,in=180] (n17);
    \draw[depedge] (n150) to[out=0,in=180] (n17);
    \draw[depedge] (n154) to[out=0,in=180] (n17);
    \draw[depedge] (n14) to[out=0,in=180] (n18);
    \draw[depedge] (n68) to[out=0,in=180] (n18);
    \draw[depedge] (n15) to[out=0,in=180] (n19);
    \draw[depedge] (n16) to[out=0,in=180] (n19);
    \draw[depedge] (n66) to[out=0,in=180] (n20);
    \draw[depedge] (n67) to[out=0,in=180] (n20);
    \draw[depedge] (n84) to[out=0,in=180] (n20);
    \draw[depedge] (n85) to[out=0,in=180] (n20);
    \draw[depedge] (n87) to[out=0,in=180] (n20);
    \draw[depedge] (n89) to[out=0,in=180] (n20);
    \draw[depedge] (n90) to[out=0,in=180] (n20);
    \draw[depedge] (n93) to[out=0,in=180] (n20);
    \draw[depedge] (n36) to[out=0,in=180] (n22);
    \draw[depedge] (n24) to[out=0,in=180] (n23);
    \draw[depedge] (n42) to[out=0,in=180] (n23);
    \draw[depedge] (n1) to[out=0,in=180] (n24);
    \draw[depedge] (n25) to[out=0,in=180] (n24);
    \draw[depedge] (n26) to[out=0,in=180] (n24);
    \draw[depedge] (n42) to[out=0,in=180] (n24);
    \draw[depedge] (n0) to[out=0,in=180] (n25);
    \draw[depedge] (n39) to[out=0,in=180] (n25);
    \draw[depedge] (n27) to[out=0,in=180] (n26);
    \draw[depedge] (n28) to[out=0,in=180] (n26);
    \draw[depedge] (n30) to[out=0,in=180] (n26);
    \draw[depedge] (n31) to[out=0,in=180] (n26);
    \draw[depedge] (n2) to[out=0,in=180] (n28);
    \draw[depedge] (n29) to[out=0,in=180] (n28);
    \draw[depedge] (n34) to[out=0,in=180] (n28);
    \draw[depedge] (n42) to[out=0,in=180] (n28);
    \draw[depedge] (n45) to[out=0,in=180] (n28);
    \draw[depedge] (n47) to[out=0,in=180] (n28);
    \draw[depedge] (n25) to[out=0,in=180] (n29);
    \draw[depedge] (n37) to[out=0,in=180] (n29);
    \draw[depedge] (n44) to[out=0,in=180] (n29);
    \draw[depedge] (n27) to[out=0,in=180] (n30);
    \draw[depedge] (n29) to[out=0,in=180] (n31);
    \draw[depedge] (n33) to[out=0,in=180] (n31);
    \draw[depedge] (n42) to[out=0,in=180] (n31);
    \draw[depedge] (n45) to[out=0,in=180] (n31);
    \draw[depedge] (n36) to[out=0,in=180] (n32);
    \draw[depedge] (n32) to[out=0,in=180] (n33);
    \draw[depedge] (n32) to[out=0,in=180] (n34);
    \draw[depedge] (n35) to[out=0,in=180] (n34);
    \draw[depedge] (n38) to[out=0,in=180] (n34);
    \draw[depedge] (n44) to[out=0,in=180] (n34);
    \draw[depedge] (n36) to[out=0,in=180] (n35);
    \draw[depedge] (n38) to[out=0,in=180] (n35);
    \draw[depedge] (n46) to[out=0,in=180] (n35);
    \draw[depedge] (n39) to[out=0,in=180] (n36);
    \draw[depedge] (n21) to[out=0,in=180] (n37);
    \draw[depedge] (n39) to[out=0,in=180] (n38);
    \draw[depedge] (n21) to[out=0,in=180] (n39);
    \draw[depedge] (n39) to[out=0,in=180] (n40);
    \draw[depedge] (n39) to[out=0,in=180] (n41);
    \draw[depedge] (n40) to[out=0,in=180] (n42);
    \draw[depedge] (n41) to[out=0,in=180] (n42);
    \draw[depedge] (n43) to[out=0,in=180] (n42);
    \draw[depedge] (n46) to[out=0,in=180] (n42);
    \draw[depedge] (n47) to[out=0,in=180] (n42);
    \draw[depedge] (n39) to[out=0,in=180] (n43);
    \draw[depedge] (n39) to[out=0,in=180] (n44);
    \draw[depedge] (n39) to[out=0,in=180] (n45);
    \draw[depedge] (n39) to[out=0,in=180] (n46);
    \draw[depedge] (n39) to[out=0,in=180] (n47);
    \draw[depedge] (n13) to[out=0,in=180] (n48);
    \draw[depedge] (n153) to[out=0,in=180] (n49);
    \draw[depedge] (n50) to[out=0,in=180] (n51);
    \draw[depedge] (n54) to[out=0,in=180] (n52);
    \draw[depedge] (n55) to[out=0,in=180] (n52);
    \draw[depedge] (n55) to[out=0,in=180] (n53);
    \draw[depedge] (n57) to[out=0,in=180] (n53);
    \draw[depedge] (n58) to[out=0,in=180] (n53);
    \draw[depedge] (n59) to[out=0,in=180] (n53);
    \draw[depedge] (n60) to[out=0,in=180] (n53);
    \draw[depedge] (n62) to[out=0,in=180] (n53);
    \draw[depedge] (n64) to[out=0,in=180] (n53);
    \draw[depedge] (n65) to[out=0,in=180] (n53);
    \draw[depedge] (n1) to[out=0,in=180] (n55);
    \draw[depedge] (n13) to[out=0,in=180] (n55);
    \draw[depedge] (n21) to[out=0,in=180] (n55);
    \draw[depedge] (n22) to[out=0,in=180] (n55);
    \draw[depedge] (n23) to[out=0,in=180] (n55);
    \draw[depedge] (n24) to[out=0,in=180] (n55);
    \draw[depedge] (n25) to[out=0,in=180] (n55);
    \draw[depedge] (n26) to[out=0,in=180] (n55);
    \draw[depedge] (n28) to[out=0,in=180] (n55);
    \draw[depedge] (n29) to[out=0,in=180] (n55);
    \draw[depedge] (n31) to[out=0,in=180] (n55);
    \draw[depedge] (n33) to[out=0,in=180] (n55);
    \draw[depedge] (n34) to[out=0,in=180] (n55);
    \draw[depedge] (n35) to[out=0,in=180] (n55);
    \draw[depedge] (n36) to[out=0,in=180] (n55);
    \draw[depedge] (n38) to[out=0,in=180] (n55);
    \draw[depedge] (n40) to[out=0,in=180] (n55);
    \draw[depedge] (n42) to[out=0,in=180] (n55);
    \draw[depedge] (n43) to[out=0,in=180] (n55);
    \draw[depedge] (n46) to[out=0,in=180] (n55);
    \draw[depedge] (n47) to[out=0,in=180] (n55);
    \draw[depedge] (n48) to[out=0,in=180] (n55);
    \draw[depedge] (n49) to[out=0,in=180] (n55);
    \draw[depedge] (n51) to[out=0,in=180] (n55);
    \draw[depedge] (n69) to[out=0,in=180] (n55);
    \draw[depedge] (n153) to[out=0,in=180] (n55);
    \draw[depedge] (n56) to[out=0,in=180] (n60);
    \draw[depedge] (n57) to[out=0,in=180] (n60);
    \draw[depedge] (n61) to[out=0,in=180] (n60);
    \draw[depedge] (n62) to[out=0,in=180] (n60);
    \draw[depedge] (n63) to[out=0,in=180] (n60);
    \draw[depedge] (n58) to[out=0,in=180] (n64);
    \draw[depedge] (n12) to[out=0,in=180] (n68);
    \draw[depedge] (n68) to[out=0,in=180] (n69);
    \draw[depedge] (n74) to[out=0,in=180] (n70);
    \draw[depedge] (n9) to[out=0,in=180] (n71);
    \draw[tgtedge] (n6) to[out=0,in=180] (n72);
    \draw[tgtedge] (n19) to[out=0,in=180] (n72);
    \draw[tgtedge] (n52) to[out=0,in=180] (n72);
    \draw[tgtedge] (n71) to[out=0,in=180] (n72);
    \draw[tgtedge] (n73) to[out=0,in=180] (n72);
    \draw[tgtedge] (n132) to[out=0,in=180] (n72);
    \draw[tgtedge] (n149) to[out=0,in=180] (n72);
    \draw[tgtedge] (n151) to[out=0,in=180] (n72);
    \draw[depedge] (n9) to[out=0,in=180] (n73);
    \draw[depedge] (n74) to[out=0,in=180] (n75);
    \draw[depedge] (n75) to[out=0,in=180] (n76);
    \draw[depedge] (n92) to[out=0,in=180] (n76);
    \draw[depedge] (n78) to[out=0,in=180] (n77);
    \draw[depedge] (n75) to[out=0,in=180] (n78);
    \draw[depedge] (n86) to[out=0,in=180] (n79);
    \draw[depedge] (n88) to[out=0,in=180] (n79);
    \draw[depedge] (n76) to[out=0,in=180] (n80);
    \draw[depedge] (n77) to[out=0,in=180] (n80);
    \draw[depedge] (n74) to[out=0,in=180] (n81);
    \draw[depedge] (n81) to[out=0,in=180] (n82);
    \draw[depedge] (n87) to[out=0,in=180] (n82);
    \draw[depedge] (n74) to[out=0,in=180] (n83);
    \draw[depedge] (n74) to[out=0,in=180] (n84);
    \draw[depedge] (n75) to[out=0,in=180] (n85);
    \draw[depedge] (n84) to[out=0,in=180] (n85);
    \draw[depedge] (n78) to[out=0,in=180] (n86);
    \draw[depedge] (n83) to[out=0,in=180] (n87);
    \draw[depedge] (n91) to[out=0,in=180] (n87);
    \draw[depedge] (n92) to[out=0,in=180] (n87);
    \draw[depedge] (n76) to[out=0,in=180] (n88);
    \draw[depedge] (n90) to[out=0,in=180] (n88);
    \draw[depedge] (n87) to[out=0,in=180] (n89);
    \draw[depedge] (n75) to[out=0,in=180] (n90);
    \draw[depedge] (n87) to[out=0,in=180] (n90);
    \draw[depedge] (n75) to[out=0,in=180] (n91);
    \draw[depedge] (n74) to[out=0,in=180] (n92);
    \draw[depedge] (n76) to[out=0,in=180] (n93);
    \draw[depedge] (n77) to[out=0,in=180] (n93);
    \draw[depedge] (n80) to[out=0,in=180] (n93);
    \draw[depedge] (n87) to[out=0,in=180] (n93);
    \draw[depedge] (n88) to[out=0,in=180] (n93);
    \draw[depedge] (n94) to[out=0,in=180] (n93);
    \draw[depedge] (n96) to[out=0,in=180] (n93);
    \draw[depedge] (n79) to[out=0,in=180] (n94);
    \draw[depedge] (n86) to[out=0,in=180] (n94);
    \draw[depedge] (n87) to[out=0,in=180] (n94);
    \draw[depedge] (n86) to[out=0,in=180] (n95);
    \draw[depedge] (n91) to[out=0,in=180] (n95);
    \draw[depedge] (n87) to[out=0,in=180] (n96);
    \draw[depedge] (n95) to[out=0,in=180] (n96);
    \draw[tgtedge] (n11) to[out=0,in=180] (n97);
    \draw[tgtedge] (n82) to[out=0,in=180] (n97);
    \draw[tgtedge] (n98) to[out=0,in=180] (n97);
    \draw[tgtedge] (n99) to[out=0,in=180] (n97);
    \draw[tgtedge] (n100) to[out=0,in=180] (n97);
    \draw[tgtedge] (n101) to[out=0,in=180] (n97);
    \draw[tgtedge] (n102) to[out=0,in=180] (n97);
    \draw[tgtedge] (n103) to[out=0,in=180] (n97);
    \draw[tgtedge] (n117) to[out=0,in=180] (n97);
    \draw[tgtedge] (n130) to[out=0,in=180] (n97);
    \draw[depedge] (n133) to[out=0,in=180] (n100);
    \draw[depedge] (n11) to[out=0,in=180] (n103);
    \draw[depedge] (n82) to[out=0,in=180] (n103);
    \draw[depedge] (n117) to[out=0,in=180] (n103);
    \draw[depedge] (n106) to[out=0,in=180] (n104);
    \draw[depedge] (n104) to[out=0,in=180] (n105);
    \draw[depedge] (n107) to[out=0,in=180] (n105);
    \draw[depedge] (n108) to[out=0,in=180] (n105);
    \draw[depedge] (n110) to[out=0,in=180] (n105);
    \draw[depedge] (n106) to[out=0,in=180] (n107);
    \draw[depedge] (n106) to[out=0,in=180] (n108);
    \draw[depedge] (n137) to[out=0,in=180] (n108);
    \draw[tgtedge] (n104) to[out=0,in=180] (n109);
    \draw[tgtedge] (n105) to[out=0,in=180] (n109);
    \draw[tgtedge] (n140) to[out=0,in=180] (n109);
    \draw[depedge] (n74) to[out=0,in=180] (n111);
    \draw[depedge] (n59) to[out=0,in=180] (n113);
    \draw[depedge] (n102) to[out=0,in=180] (n114);
    \draw[tgtedge] (n11) to[out=0,in=180] (n115);
    \draw[tgtedge] (n99) to[out=0,in=180] (n115);
    \draw[tgtedge] (n102) to[out=0,in=180] (n115);
    \draw[tgtedge] (n103) to[out=0,in=180] (n115);
    \draw[tgtedge] (n112) to[out=0,in=180] (n115);
    \draw[tgtedge] (n113) to[out=0,in=180] (n115);
    \draw[tgtedge] (n114) to[out=0,in=180] (n115);
    \draw[tgtedge] (n117) to[out=0,in=180] (n115);
    \draw[tgtedge] (n130) to[out=0,in=180] (n115);
    \draw[depedge] (n74) to[out=0,in=180] (n116);
    \draw[depedge] (n87) to[out=0,in=180] (n117);
    \draw[depedge] (n116) to[out=0,in=180] (n117);
    \draw[depedge] (n111) to[out=0,in=180] (n118);
    \draw[depedge] (n116) to[out=0,in=180] (n118);
    \draw[depedge] (n12) to[out=0,in=180] (n119);
    \draw[depedge] (n128) to[out=0,in=180] (n119);
    \draw[depedge] (n127) to[out=0,in=180] (n120);
    \draw[depedge] (n120) to[out=0,in=180] (n121);
    \draw[depedge] (n129) to[out=0,in=180] (n121);
    \draw[depedge] (n120) to[out=0,in=180] (n122);
    \draw[depedge] (n125) to[out=0,in=180] (n122);
    \draw[depedge] (n69) to[out=0,in=180] (n123);
    \draw[depedge] (n70) to[out=0,in=180] (n123);
    \draw[depedge] (n83) to[out=0,in=180] (n123);
    \draw[depedge] (n119) to[out=0,in=180] (n123);
    \draw[depedge] (n124) to[out=0,in=180] (n123);
    \draw[depedge] (n126) to[out=0,in=180] (n123);
    \draw[depedge] (n127) to[out=0,in=180] (n123);
    \draw[depedge] (n128) to[out=0,in=180] (n123);
    \draw[depedge] (n129) to[out=0,in=180] (n123);
    \draw[depedge] (n153) to[out=0,in=180] (n123);
    \draw[depedge] (n0) to[out=0,in=180] (n124);
    \draw[depedge] (n2) to[out=0,in=180] (n124);
    \draw[depedge] (n120) to[out=0,in=180] (n124);
    \draw[depedge] (n121) to[out=0,in=180] (n124);
    \draw[depedge] (n122) to[out=0,in=180] (n124);
    \draw[depedge] (n125) to[out=0,in=180] (n124);
    \draw[depedge] (n129) to[out=0,in=180] (n124);
    \draw[depedge] (n152) to[out=0,in=180] (n124);
    \draw[depedge] (n152) to[out=0,in=180] (n125);
    \draw[depedge] (n70) to[out=0,in=180] (n126);
    \draw[depedge] (n119) to[out=0,in=180] (n126);
    \draw[depedge] (n153) to[out=0,in=180] (n126);
    \draw[depedge] (n68) to[out=0,in=180] (n127);
    \draw[depedge] (n84) to[out=0,in=180] (n127);
    \draw[depedge] (n119) to[out=0,in=180] (n127);
    \draw[depedge] (n74) to[out=0,in=180] (n128);
    \draw[depedge] (n87) to[out=0,in=180] (n129);
    \draw[depedge] (n128) to[out=0,in=180] (n129);
    \draw[depedge] (n19) to[out=0,in=180] (n130);
    \draw[depedge] (n123) to[out=0,in=180] (n130);
    \draw[depedge] (n149) to[out=0,in=180] (n130);
    \draw[depedge] (n134) to[out=0,in=180] (n131);
    \draw[depedge] (n131) to[out=0,in=180] (n132);
    \draw[depedge] (n135) to[out=0,in=180] (n132);
    \draw[depedge] (n136) to[out=0,in=180] (n132);
    \draw[depedge] (n137) to[out=0,in=180] (n132);
    \draw[depedge] (n131) to[out=0,in=180] (n133);
    \draw[depedge] (n137) to[out=0,in=180] (n133);
    \draw[depedge] (n134) to[out=0,in=180] (n135);
    \draw[depedge] (n134) to[out=0,in=180] (n136);
    \draw[depedge] (n138) to[out=0,in=180] (n137);
    \draw[depedge] (n134) to[out=0,in=180] (n138);
    \draw[depedge] (n12) to[out=0,in=180] (n139);
    \draw[depedge] (n53) to[out=0,in=180] (n140);
    \draw[depedge] (n110) to[out=0,in=180] (n140);
    \draw[depedge] (n139) to[out=0,in=180] (n140);
    \draw[depedge] (n141) to[out=0,in=180] (n140);
    \draw[depedge] (n142) to[out=0,in=180] (n140);
    \draw[depedge] (n143) to[out=0,in=180] (n140);
    \draw[depedge] (n145) to[out=0,in=180] (n140);
    \draw[depedge] (n146) to[out=0,in=180] (n140);
    \draw[depedge] (n147) to[out=0,in=180] (n140);
    \draw[depedge] (n148) to[out=0,in=180] (n140);
    \draw[depedge] (n53) to[out=0,in=180] (n141);
    \draw[depedge] (n142) to[out=0,in=180] (n143);
    \draw[depedge] (n144) to[out=0,in=180] (n143);
    \draw[depedge] (n69) to[out=0,in=180] (n144);
    \draw[depedge] (n146) to[out=0,in=180] (n144);
    \draw[depedge] (n146) to[out=0,in=180] (n145);
    \draw[depedge] (n68) to[out=0,in=180] (n146);
    \draw[depedge] (n139) to[out=0,in=180] (n146);
    \draw[depedge] (n139) to[out=0,in=180] (n147);
    \draw[depedge] (n13) to[out=0,in=180] (n148);
    \draw[depedge] (n139) to[out=0,in=180] (n148);
    \draw[depedge] (n153) to[out=0,in=180] (n148);
    \draw[depedge] (n131) to[out=0,in=180] (n149);
    \draw[depedge] (n132) to[out=0,in=180] (n149);
    \draw[depedge] (n135) to[out=0,in=180] (n149);
    \draw[depedge] (n136) to[out=0,in=180] (n149);
    \draw[depedge] (n137) to[out=0,in=180] (n149);
    \draw[depedge] (n150) to[out=0,in=180] (n149);
    \draw[depedge] (n150) to[out=0,in=180] (n151);
    \draw[depedge] (n20) to[out=0,in=180] (n152);
    \draw[depedge] (n12) to[out=0,in=180] (n153);
  \end{pgfonlayer}
\end{tikzpicture}
}
\caption{The terminal proof tablet ($156$ nodes), laid left-to-right by import depth; edges are direct \lean{} imports. Squares are definitions, circles theorem-like nodes, labeled red pills the five paper targets; boldly drawn nodes are the \ncoarse{} \emph{coarse} nodes fixed at the phase boundary, the faded ones the fine scaffolding added during formalization. The four enlarged red squares are the targets' kernel-computed \emph{semantic closure}---the only nodes the targets' meaning depends on, and thus the human-protected core: of all $156$ nodes, only edits to these four (plus the five targets) can change what the targets mean.}
\label{fig:tablet}
\end{figure*}

\tightparagraph{The shape of the finished artifact.} Figure~\ref{fig:tablet} renders the terminal tablet as the kernel sees it: a single DAG whose deepest import chains run $17$ steps from a primitive definition up to a paper target. Figure~\ref{fig:targetcore} shows the meaning-bearing root of that DAG concretely---the natural-language/\lean{} pair for the main theorem and for the two definitions in its semantic closure, which are exactly the shapes a human ratifies before formalization proceeds. The structure is not a tree of five independent proofs but a shared lattice---$119$ of the $156$ nodes lie in more than one target's cone, and $10$ foundational nodes lie in all five---which mirrors the paper itself, where one pseudorandom construction underlies every headline bound. 

\usetikzlibrary{fit,calc,backgrounds,positioning,arrows.meta,shapes.geometric}

\tikzset{
  psBlane/.style   ={rounded corners=2pt,draw=clane!55!black,line width=0.5pt,
                     fill=clane!7,inner sep=0pt},
  psBphase/.style  ={rounded corners=3pt,draw=cphase,line width=0.7pt,
                     fill=cphase!8,inner sep=0pt},
  psBact/.style    ={rectangle,rounded corners=2pt,draw=clane!60!black,
                     line width=0.5pt,fill=clane!12,align=center,
                     font=\sffamily\fontsize{6.4}{7.2}\selectfont,
                     inner sep=2.4pt,minimum height=8pt},
  psBkernel/.style ={rectangle,rounded corners=2pt,draw=black!70,line width=0.6pt,
                     fill=black!7,align=center,
                     font=\sffamily\fontsize{6.4}{7.2}\selectfont,
                     inner sep=2.4pt,minimum height=8pt},
  psBver/.style    ={rectangle,rounded corners=2pt,draw=cthmnode!60!black,
                     line width=0.5pt,fill=cthm!22,align=center,
                     font=\sffamily\fontsize{6.2}{7}\selectfont,
                     inner sep=2.2pt,minimum height=8pt},
  psBaudit/.style  ={rectangle,rounded corners=2pt,draw=cred!60!black,
                     line width=0.5pt,fill=credfill!30,align=center,
                     font=\sffamily\fontsize{6.2}{7}\selectfont,
                     inner sep=2.2pt,minimum height=8pt},
  psBcheck/.style  ={rectangle,rounded corners=2pt,draw=cgreen!55!black,
                     line width=0.5pt,fill=cgreen!14,align=center,
                     font=\sffamily\fontsize{6.2}{7}\selectfont,
                     inner sep=2.2pt,minimum height=8pt},
  psBconsult/.style={draw=cgreen!60!black,double,double distance=1.6pt,
                     line width=0.5pt},
  psBgate/.style   ={diamond,aspect=2.2,draw=ctgt,line width=0.7pt,fill=ctgt!10,
                     text=ctgtlabeltext,align=center,
                     font=\sffamily\bfseries\fontsize{6}{6.6}\selectfont,
                     inner sep=1pt},
  psBstate/.style  ={rectangle,rounded corners=2pt,draw=clane!55!black,
                     line width=0.5pt,fill=clane!10,align=center,
                     font=\ttfamily\fontsize{6.2}{7}\selectfont,inner sep=2.4pt},
  psBflow/.style   ={-{Stealth[length=3.4pt]},draw=black!72,line width=0.55pt},
  psBflowd/.style  ={-{Stealth[length=3.4pt]},draw=black!55,line width=0.5pt,
                     dashed,dash pattern=on 1.6pt off 1.4pt},
  psBreopen/.style ={-{Stealth[length=3.4pt]},draw=cred,line width=0.8pt},
  psBlbl/.style    ={font=\sffamily\fontsize{5.7}{6.4}\selectfont,
                     text=black!72,inner sep=1pt},
  psBhead/.style   ={font=\sffamily\bfseries\fontsize{6.6}{7.6}\selectfont},
  psBtitle/.style  ={font=\sffamily\bfseries\fontsize{7.6}{8.6}\selectfont},
}

\newcommand{\circref}[1]{\textnormal{(#1)}}

\section{Details of the Trellis design}
\label{sec:process-semantics}
Our Trellis implementation sits in a kernel of (at time of writing) 35k lines of Rust code, supported by another 20k lines of python plumbing for agent calls and so forth.  Our source repo at \cite{source} also includes a detailed \tla{}~\cite{tlaplus} spec of the process of 10k lines.  In this section we aim to clearly describe a simplified view of the most salient details of process semantics and implementation.

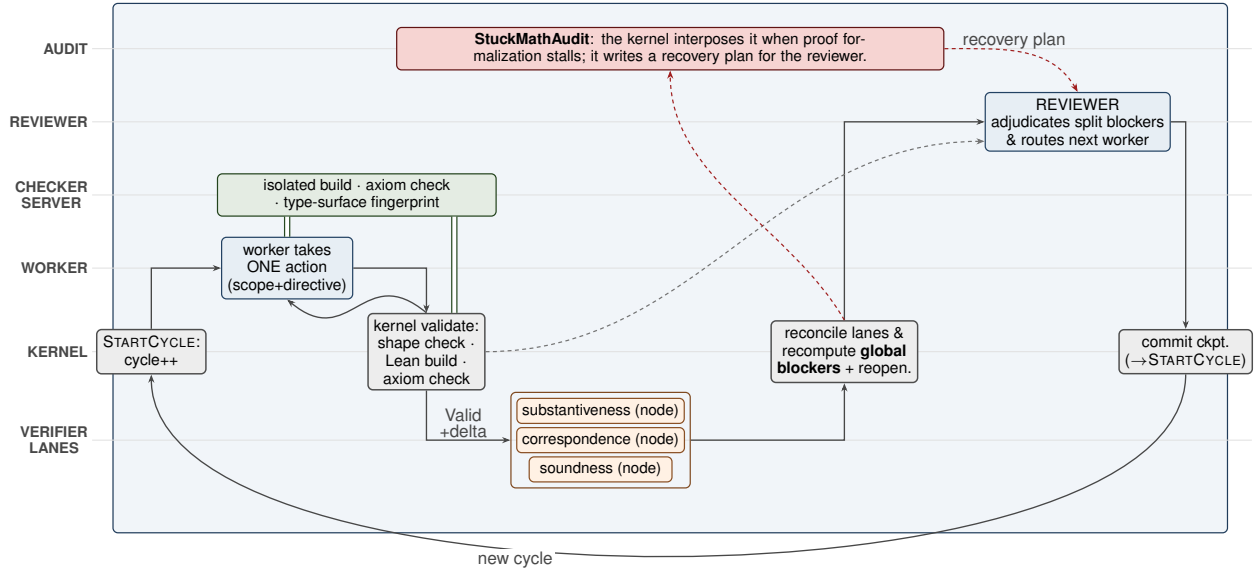
\begin{figure*}[tbp]
\centering
\setlength{\fboxsep}{0pt}
\resizebox{\textwidth}{!}{%
\begin{tikzpicture}[font=\sffamily]
\node[psBlane,minimum width=16.0cm,minimum height=7.6cm] (Bbox) at (0,0.15) {};

\def\Lx{-8.3}
\def\Rx{8.35}
\foreach \y/\name in {
   3.3/{AUDIT},
   2.25/{REVIEWER},
   1.2/{CHECKER\\SERVER},
   0.15/{WORKER},
   -1.05/{KERNEL},
   -2.32/{VERIFIER\\LANES}}{
  \node[psBlbl,anchor=east,align=center,font=\sffamily\bfseries\fontsize{6}{6.6}\selectfont]
     at (\Lx-0.02,\y) {\name};
  \draw[draw=black!12,line width=0.4pt] (\Lx,\y) -- (\Rx,\y);
}

\node[psBkernel] (Bstart) at (-7.45,-1.05)
  {\textsc{StartCycle}:\\cycle++};
\node[psBkernel] (Bval) at (-3.5,-1.05)
  {kernel validate:\\shape check $\cdot$\\\lean{} build $\cdot$\\axiom check};
\node[psBkernel] (Bblock) at (2.5,-1.05)
  {reconcile lanes \&\\recompute \textbf{global}\\\textbf{blockers} + reopen.};
\node[psBkernel] (Bcommit) at (7.4,-1.05)
  {commit ckpt.\\($\to$\textsc{StartCycle})};

\node[psBact] (Bwork) at (-5.5,0.15) {worker takes\\ONE action\\(scope+directive)};

\node[psBact] (Brev) at (5.85,2.25)
  {REVIEWER\\adjudicates split blockers\\\& routes next worker};

\node[psBcheck,minimum width=4.0cm] (Bchk) at (-4.5,1.2)
  {isolated build $\cdot$ axiom check\\$\cdot$ type-surface fingerprint};

\node[psBver,minimum width=2.05cm] (Bsubst) at (-1.0,-1.9)  {substantiveness (node)};
\node[psBver,minimum width=2.05cm] (Bcorr)  at (-1.0,-2.32) {correspondence (node)};
\node[psBver,minimum width=2.05cm] (Bsound) at (-1.0,-2.74) {soundness (node)};
\node[draw=cthmnode!55!black,line width=0.5pt,rounded corners=2pt,
      fit=(Bsubst)(Bsound),inner sep=2.2pt] (Bpanel) {};

\node[psBaudit,align=center,text width=7.7cm] (Bsma) at (0,3.3)
  {\textbf{StuckMathAudit}: the kernel interposes it when proof formalization stalls; it writes a recovery plan for the reviewer.};

\draw[psBflow] (Bstart.north) |- (Bwork.west);
\draw[psBflow] (Bwork.east) -| (Bval.north);
\draw[psBflow] (Bval.south) |- node[psBlbl,above,pos=0.72,align=center]{\scriptsize Valid\\\scriptsize +delta} (Bpanel.west);
\draw[psBflow] (Bpanel.east) -| (Bblock.south);
\draw[psBflow] (Bblock.north) |- (Brev.west);
\draw[psBflow] (Brev.east) -| (Bcommit.north);
\draw[psBconsult] (Bchk.south -| Bwork.north) -- (Bwork.north);
\draw[psBconsult] ([xshift=4mm]Bval.north |- Bchk.south) -- ([xshift=4mm]Bval.north);
\draw[psBflowd] (Bval.east) to[out=0,in=180]
  ($(Brev.west)+(0,-0.28)$);
\draw[psBflow] (Bval.north) to[out=140,in=-40,looseness=1.5] (Bwork.south);
\draw[psBflow] (Bcommit.south) to[out=-90,in=-90,looseness=0.6]
  node[psBlbl,below,pos=0.6,fill=white,inner sep=1pt]{\scriptsize new cycle} (Bstart.south);
\draw[psBflowd,draw=cred] (Bblock.north) to[out=120,in=-90,looseness=0.9]
  node[psBlbl,left,pos=0.55]{} (Bsma.south);
\draw[psBflowd,draw=cred] (Bsma.east) to[out=0,in=120]
  node[psBlbl,above,pos=0.45]{\scriptsize recovery plan} (Brev.north);

\end{tikzpicture}%
}
\caption{One supervisor cycle in proof formalization. The kernel issues exactly one agent call at a time; time runs left to right and each swimlane is the actor the kernel consults. Solid arrows are the normal control flow; the black dashed arrow carries outcomes that bypass the verifier lanes straight to the reviewer; the arrow returning to the worker is an \texttt{Invalid} retry. A green double line marks each step that consults the isolated checker server, which has its own lane: the worker test-builds its edits there, and the kernel's validation runs the authoritative build, axiom check, and type-surface fingerprint there. The red dashed detour in the top lane is the occasional StuckMathAudit agent, which the kernel interposes when formalization stalls and which only advises the reviewer. Of the agents, only the worker can write to the Tablet.}
\label{fig:ps-cycle}
\end{figure*}

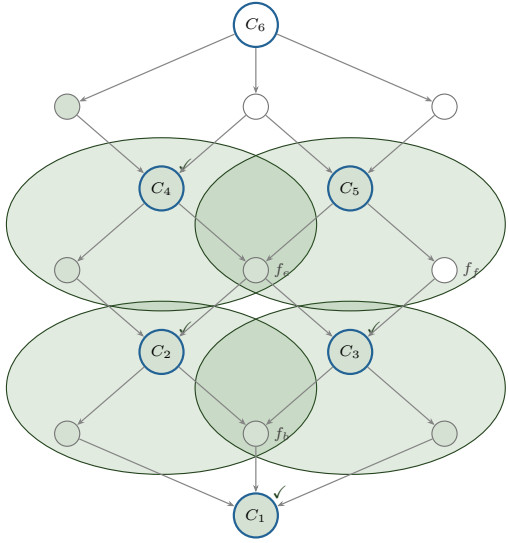
\begin{figure}[t]
\centering
\resizebox{0.82\linewidth}{!}{%
\begin{tikzpicture}[
  font=\sffamily,
  coarse/.style={circle,minimum size=7mm,line width=1.0pt,draw=cdefnode!80!black,inner sep=0pt,font=\fontsize{6.6}{7.5}\selectfont},
  fine/.style={circle,minimum size=4mm,line width=0.5pt,draw=black!55,inner sep=0pt},
  cls/.style={fill=cgreen!22},
  opn/.style={fill=white},
  ed/.style={-{Stealth[length=3.4pt]},draw=black!48,line width=0.55pt},
  scc/.style={text=cgreen!50!black,font=\footnotesize\bfseries},
  lbl/.style={font=\fontsize{7}{8.5}\selectfont,text=black!75},
  clo/.style={rounded corners=4pt,draw=cgreen!55!black,line width=0.5pt,fill=cgreen,fill opacity=0.15},
]
\node[coarse,cls] (C1) at (0,0) {$C_1$};
\node[fine,cls] (fa) at (-3,1.3) {};
\node[fine,cls] (fb) at (0,1.3) {};
\node[fine,cls] (fc) at (3,1.3) {};
\node[coarse,cls] (C2) at (-1.5,2.6) {$C_2$};
\node[coarse,cls] (C3) at (1.5,2.6) {$C_3$};
\node[fine,cls] (fd) at (-3,3.9) {};
\node[fine,cls] (fe) at (0,3.9) {};
\node[fine,opn] (ff) at (3,3.9) {};
\node[coarse,cls] (C4) at (-1.5,5.2) {$C_4$};
\node[coarse,cls] (C5) at (1.5,5.2) {$C_5$};
\node[fine,cls] (fg) at (-3,6.5) {};
\node[fine,opn] (fh) at (0,6.5) {};
\node[fine,opn] (fi) at (3,6.5) {};
\node[coarse,opn] (C6) at (0,7.8) {$C_6$};
\node[lbl,font=\scriptsize\itshape,anchor=west,inner sep=2pt] at (fb.east) {$f_b$};
\node[lbl,font=\scriptsize\itshape,anchor=west,inner sep=2pt] at (fe.east) {$f_e$};
\node[lbl,font=\scriptsize\itshape,anchor=west,inner sep=2pt] at (ff.east) {$f_f$};
\foreach \n in {C1,C2,C3,C4} {\node[scc] at ([shift={(1.2mm,1.2mm)}]\n.north east) {\checkmark};}
\begin{pgfonlayer}{background}
  \node[ellipse,clo,fit=(C2)(fa)(fb),inner sep=1pt] {};
  \node[ellipse,clo,fit=(C3)(fc)(fb),inner sep=1pt] {};
  \node[ellipse,clo,fit=(C4)(fd)(fe),inner sep=1pt] {};
  \node[ellipse,clo,fit=(C5)(ff)(fe),inner sep=1pt] {};
\end{pgfonlayer}
\draw[ed] (fa)--(C1); \draw[ed] (fb)--(C1); \draw[ed] (fc)--(C1);
\draw[ed] (C2)--(fa); \draw[ed] (C2)--(fb);
\draw[ed] (C3)--(fb); \draw[ed] (C3)--(fc);
\draw[ed] (fd)--(C2); \draw[ed] (fe)--(C2); \draw[ed] (fe)--(C3); \draw[ed] (ff)--(C3);
\draw[ed] (C4)--(fd); \draw[ed] (C4)--(fe);
\draw[ed] (C5)--(fe); \draw[ed] (C5)--(ff);
\draw[ed] (fg)--(C4); \draw[ed] (fh)--(C4); \draw[ed] (fh)--(C5); \draw[ed] (fi)--(C5);
\draw[ed] (C6)--(fg); \draw[ed] (C6)--(fh); \draw[ed] (C6)--(fi);
\end{tikzpicture}%
}
\caption{\word{Coarse focus.} Large nodes are \word{coarse} (already present when the proof formalization phase begins), small ones \word{fine} (added during proof formalization); green $=$ \lean{}-closed, white $=$ open. A coarse node is \word{shallow-coarse-closed} (\checkmark) when it and everything reachable along import edges (pointing downward, toward dependencies) up to the next coarse node is closed. Shaded regions are such closures; they overlap at shared fine nodes ($f_b$, $f_e$). $C_5$ is \lean{}-closed but not shallow-coarse-closed, since $f_f$ is open.}
\label{fig:ps-coarse}
\end{figure}

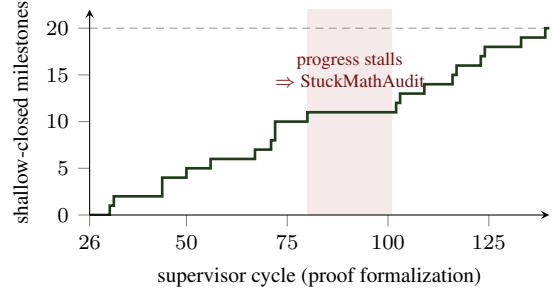
\begin{figure}[t]
\centering
\begin{tikzpicture}
\begin{axis}[
  width=0.95\linewidth, height=4.3cm,
  xmin=\coarsepfstart, xmax=140, ymin=0, ymax=22,
  enlarge x limits=false,
  xlabel={supervisor cycle (proof formalization)},
  ylabel={shallow-closed milestones},
  xlabel style={font=\footnotesize}, ylabel style={font=\footnotesize},
  tick label style={font=\footnotesize},
  ytick={0,5,10,15,20}, xtick={26,50,75,100,125},
  axis lines=left, tick align=outside, axis on top,
]
  \fill[cred!10] (axis cs:\coarseplatstart,0) rectangle (axis cs:\coarseplatend,22);
  \draw[densely dashed,draw=black!40] (axis cs:\coarsepfstart,\coarsetotal) -- (axis cs:140,\coarsetotal);
  \addplot[const plot,line width=1pt,draw=cgreen!50!black] table[x=cycle,y=closed] {figs/coarse.dat};
  \node[font=\scriptsize,text=cred!72!black,align=center,anchor=south] at (axis cs:90.5,12.7) {progress stalls\\$\Rightarrow$ StuckMathAudit};
\end{axis}
\end{tikzpicture}
\caption{Shallow-coarse progress in the run on the Brada\v{c} paper: how many of the $\coarsetotal{}$ coarse milestones are fully shallow-coarse-closed, by cycle. It rises monotonically to all $\coarsetotal{}$; the shaded plateau (cycles~$\coarseplatstart$--$\coarseplatend$) is a stall that tripped a StuckMathAudit.}
\label{fig:ps-coarse-progress}
\end{figure}

\subsection{The Tablet}
In a Trellis autoformalization run, the \word{Tablet} of \lean{}/\LaTeX{} \word{nodes} is itself a git repository.  Among all agents in the process, only the worker agent has write access to the Tablet (the deterministic supervisor does as well), and only the deterministic supervisor has write access to the .git directory controlling history.  Worker edits accepted by the deterministic supervisor are committed as \word{checkpoints} (git commits) to the Tablet git repository; rejected edits are rolled back.  We refer to any node failing one of the verifier lanes as a \word{blocker}; a \word{clean checkpoint} is a git checkpoint with no blockers.  The initial \word{theorem stating} phase is the phase where the process aims to reach a clean checkpoint.  After that, in \word{proof formalization}, the process aims to close all \lean{} proofs with limited drift from clean checkpoints.  That is, work may involve changes to nodes that temporarily invalidate verifications of other nodes, but we expect to be able to repair these issues and get back to a clean checkpoint, otherwise the reviewer can choose (or be forced to choose) \word{LastClean}, a git rewind of the tablet to the last clean checkpoint.

\subsection{The supervisor cycle}
Within a phase the supervisor advances the run in \word{cycles}, the pipeline of Figure~\ref{fig:ps-cycle}. The \word{supervisor} is the process that drives the deterministic side of the run; at its core is a \word{kernel}---a state machine, written in Rust and specified in \tla{}, that owns all protocol state and makes every decision (the \textsc{kernel} row of the figure)---which the supervisor runs while also performing the I/O the kernel abstracts away: dispatching each stochastic agent in a sandbox, and running the authoritative \lean{} build through an isolated checker the agents cannot reach. The remaining rows of the figure are the agents the supervisor consults. The supervisor keeps exactly one agent \word{request} in flight at any moment and commits a checkpoint only at the close of the cycle. A cycle begins by choosing an active node together with a \word{scope}---the set of nodes the worker is permitted to edit---and a \word{directive} fixing the kind of edit expected, such as closing the active node's own proof or restructuring the support beneath it. The worker returns a single edit to the Tablet, and the supervisor checks it deterministically before any other agent sees it: a \word{shape check} that the touched files have the required form (one principal \lean{} declaration named for the node, a single \LaTeX{} statement block, and dependencies cited both by \lean{} \texttt{import} and by \verb|\noderef|); an \word{isolated build} in which the \lean{} sources are recompiled from scratch in isolation, against a workspace the worker cannot tamper with; and an \word{axiom check} that the build rests on nothing beyond a small approved axiom set, with no \texttt{sorry} hidden in a node claimed closed.  Isolated builds check whether edited Tablet nodes build correctly conditioned on the truth of their imported nodes (e.g., if the imported node statements were considered axioms).  In this way, we can work locally but be finished globally once all local work is completed.

The deterministic checks above are run in an isolated \word{checker}: a long-running, supervisor-owned server reached over a local unix socket. The checker derives the build target from its own runtime root and always compiles the \emph{supervisor's} authoritative workspace, into which the worker's edited Tablet sources are first mirrored by a hardened, fingerprint-cached sync. The same checker is consulted by two clients of different standing: the untrusted worker calls it during its turn to test-build and iterate on its own edits, while the kernel calls it for the authoritative validation of a submitted edit. Because both routes compile the identical supervisor workspace, the worker's use is a convenience that cannot move the verdict; the kernel's call is the binding one. The checker performs the build-bearing checks---the isolated build of a node, the axiom check, and the \lean{} type-surface fingerprint---while the purely structural shape check and the approved-axiom \emph{policy} decision remain in the kernel, which only consumes the checker's reported facts. The conditional character of the build is exactly as above: when a node's local closure reaches an imported Tablet theorem, the checker records it as a boundary helper assumed \emph{at its stated type} rather than re-elaborating its proof, and the kernel verifies that boundary statement against the imported node's ratified statement, so that node-local checks compose into the global guarantee.  An important design principle motivates the use of the checker server: we want isolated, independent checks of the worker's work, but we also want the worker to have access to exactly the same checks that will be used to accept or reject the worker's delta, so that cycles are not wasted on worker cycles that were genuinely productive but rejected because of differences between worker and supervisor build environments or assumptions.

\subsection{Verification and fingerprints}
An edit that changes Tablet content carries a \word{semantic delta}. When there is a delta, the supervisor dispatches the changed nodes to the \word{verifier lanes}, independent agents that each certify one semantic relation. The three per-node lanes are \word{substantiveness}---the node's \LaTeX{} statement is a claim the paper genuinely uses, at the paper's strength, and is not in essence a restatement of another node; \word{correspondence}---the node's \lean{} signature captures the full, precise meaning of its \LaTeX{} statement; and \word{soundness}---the node's \LaTeX{} proof is a line-by-line checkable argument from the statements of the nodes it cites. An approving verdict is stored with an \word{approved fingerprint}, a content hash of exactly the material relevant to the lane's judgment (Table~\ref{tab:ps-fingerprint}); the supervisor keeps each node's current fingerprint beside it and, whenever an edit elsewhere makes the two diverge, it automatically \word{reopens} that lane, returning the node to the blocker set.\footnote{Downstream reopening of this kind is what maintains correspondence, soundness, and---per target---paper-faithfulness, whose fingerprints draw in the imported type-surfaces, cited statements, and covering sets that a change elsewhere can disturb; keeping these current is what holds the DAG formalizable and faithful to the paper. Substantiveness is a deliberate exception: its fingerprint is node-local---a node's own \LaTeX{} statement judged against the paper---so a downstream edit does not reopen it. It serves chiefly as a guard against unproductive decomposition (Clause~2 of \S\ref{sec:gates}) rather than a relation the rest of the DAG must preserve, and reopening it on downstream changes, though a natural alternative, has in our experience seemed unnecessary.} Having run the lanes and recomputed every such reopening, the supervisor rebuilds the global blocker set and hands the cycle to the \word{reviewer}, the single adjudicator that resolves each node still \word{Unknown} and chooses where the worker goes next.

What the correspondence fingerprint hashes from the \lean{} side is not the whole declaration but its \word{type-surface}: the part that fixes the statement's mathematical meaning. The supervisor computes it by walking outward from the node's principal declaration through the constants its \emph{type} mentions, transitively, keeping what determines meaning and dropping what does not. From a theorem or lemma it takes the statement (its \texttt{type}) but never the proof term, so a lemma used only inside another node's proof never enters that node's type-surface; from a definition it takes both the type and the \emph{value}, since a definition's value is part of its meaning ($D:=3$ becoming $D:=4$ must reopen everything that transitively uses $D$); from an inductive it takes the type together with its constructors. The walk stops at the Mathlib boundary: a referenced constant defined outside the Tablet is recorded by name but not unfolded, with shifts in the external library caught instead by the pinned toolchain and \texttt{lake}-manifest hashes. The collected surface is then hashed---canonicalizing away source-position metadata and bound-variable names, but not implicit-versus-explicit binder information---to give the \lean{} component of the correspondence fingerprint. Because proof terms never enter, rewriting or reorganizing a \lean{} proof changes no correspondence fingerprint; only a change to a statement's type or a definition's value, on the node or on something its statement transitively uses, reopens correspondence.

This same type-surface walk defines the paper targets' human-protected \word{semantic closure} (Figure~\ref{fig:targetcore}): the set of Tablet declarations reached by walking outward from each target declaration---taking statement types but never proof terms, definition values, and inductive constructors, and stopping at the Mathlib boundary, whose constants are treated as pinned (by the toolchain and \texttt{lake}-manifest hashes). This closure is the trusted base of the formalization, human-approved as a frozen package at the phase transition. The kernel will not mark a run \word{Complete} unless every paper target and every declaration in this closure stands in its approved state with no open blocker; and any accepted edit that changes one of these declarations---its signature, a definition's value, or which declarations its type-surface draws into the closure---alters that node's fingerprint and deterministically reopens the human-approval gate. No reviewer decision or later worker edit can route around this gate, the isolated \lean{} build, or the axiom check, so the faithfulness of the result depends only on this human-checked closure and never on the reliability of any agent.

\subsection{Authorized deviations}
\label{sec:deviations}
The substantiveness lane requires every node's statement to be a claim the paper genuinely uses, which by default forces the formalization onto the paper's exact path. Occasionally, however, it is natural to allow a node's statement to differ mathematically from the paper's path in some way (e.g., because of an error in the paper, or because formalization favors a slightly different route); a constant changed in a proof, a strengthened hypothesis, a weakened conclusion, or an alternate intermediate statement that a later step absorbs. \system{} admits such a departure only via an explicit, authorized, durable \word{deviation} artifact. A deviation lives in a single \LaTeX{}-only file under \texttt{reference/} that must state the departure, name the nodes it affects, and give a rigorous \emph{return-to-faithful} argument: a concrete account of where the difference is absorbed so that the formalization rejoins a paper-faithful step. Each deviation must be self-contained.

Authorization is decided by a fifth verifier lane, distinct from the four of \S\ref{sec:gates}. Where substantiveness, correspondence, and soundness are \emph{per-node} and paper-faithfulness is \emph{per-target}, the \word{deviation} lane is \emph{per-deviation-file}: an independent agent reads the one reference file together with the relevant paper context and DAG nodes and decides whether that single departure is authorized---a paper-bound check the kernel tracks under its own blocker (kernel \word{Deviation} kind and object), running within the paper-verification stage ahead of per-node substantiveness. The tie to node judgment is what gives a deviation force. A node whose statement departs from the paper passes substantiveness only by \emph{claiming} the deviation that licenses it, and substantiveness then judges the node, the paper, and its claimed deviations together; a claim on an unauthorized deviation does not satisfy the lane. To introduce such a node the worker writes the reference file, registers the deviation, and records the claim on each affected node in one move; to retire one it deletes the file and drops the claims. Deviation files carry their own fingerprints, so editing one reopens its authorization exactly as editing a node's content reopens its per-node lanes---the worker cannot quietly alter an authorized departure and keep the verdict.

The importance of this system is that it provides durable kernel-tracked authorizations to deviate from the paper; in the presence even of minor mathematical typos, a system like Trellis which aims to ground formalization on a paper-faithful path could struggle with inconsistent or contradictory decisions by agents tasked with that grounding.
\subsection{The reviewer and audits}
The cycle then closes at the reviewer. Normally the supervisor \word{commits} the new state as a checkpoint and the next cycle begins; but if the worker's edit fails validation---an \word{Invalid} outcome---the supervisor discards it and re-issues the same request as a fresh \word{attempt}, retrying a bounded number of times within the cycle before escalating to the reviewer regardless. Three other outcomes bypass verification and reach the reviewer directly (the dashed path in Figure~\ref{fig:ps-cycle}): the valid no-delta edit above, an action the worker abandons as \word{Stuck} because it cannot proceed under the current scope, and an action it returns as \word{NeedsRestructure}, declaring the present decomposition itself wrong and asking for broader scope. The top lane holds a read-only \word{audit} role the kernel interposes only under specific conditions, never on an ordinary cycle. \word{StuckMathAudit} is triggered at signs of stagnation; e.g. when too many cycles pass without a clean checkpoint while a soundness or substantiveness blocker remains open; it studies the stalled region and writes a recovery plan for the reviewer. Like all other non-worker agents, it does not edit the Tablet directly; it can only advise the reviewer and/or call for a git rewind of the Tablet to a previous state.

\subsection{Scope and coarse focus}
Each worker burst is granted a bounded \word{scope}: the set of already-present nodes it may edit that cycle. The reviewer selects a \word{mode}, and the kernel both bounds and enforces the choice (Table~\ref{tab:ps-scope}). In theorem stating the modes are \word{global}, which authorizes all nodes, and \word{targeted}, which authorizes only the bidirectional dependency closure---the \word{impact region}---of a chosen node. In proof formalization the modes form a ladder of increasing authority: \word{local} permits editing only the active node's own \lean{} proof together with new helper nodes in its support cone; \word{restructure} permits coordinated edits across an explicitly enumerated subset of the focus node's impact region, including signature changes to helpers introduced during formalization; and \word{coarse-restructure} additionally permits signature changes to the protected coarse nodes fixed at the phase boundary, and is the only mode under which a node's human-approved correspondence may be reopened. Throughout, the kernel distinguishes the broad \word{scope envelope} a mode \emph{could} authorize from the explicit \word{authorized nodes} the reviewer actually hands the worker---which for the restructuring modes must be a non-empty subset of the envelope, and which never implicitly includes the active node, since that node is a scope anchor and not itself edit permission. Before dispatch the kernel rejects any out-of-envelope or out-of-cone authorization, an empty authorization under a restructuring mode or a non-empty one under local, and any blocker assigned to a worker whose scope cannot reach it; after dispatch it rejects any burst whose file delta touches a node outside its authorized set.

Scope is constrained further so that the run advances monotonically (Figure~\ref{fig:ps-coarse}). A distinguished subset of nodes---the \word{coarse DAG}, the set of all nodes present when the run enters proof formalization---marks the principal milestones, the remaining nodes being the fine scaffolding beneath them. For each coarse node the kernel computes its \word{shallow-coarse-closure}: the node is shallowly closed when it, and every dependency reachable along fine import edges, is present and closed, where the walk \emph{halts at any other coarse node} rather than descending through it. One milestone's closure is thus independent of the milestones beneath it. A single coarse node is the \word{active anchor}: it confines every focus and authorization that cycle to its support cone, and is \word{locked}---the reviewer cannot advance it---until it is shallowly closed and free of blockers. The kernel's progress measure is the number of shallowly-closed coarse nodes against the committed state, and it tracks how many checkpoints have passed since that count last rose. The utility of this progress measure is that new coarse nodes are never introduced by refinement, so enforcing monotone progress with respect to this measure is feasible. When the count stalls beyond a fixed threshold the kernel interposes a \word{StuckMathAudit} in place of the next reviewer turn. A stronger \word{no-regression} rule records the coarse nodes ever observed shallowly closed and forbids advancing the active anchor while any of them has lost closure, pinning work on the regressed material until it recovers; after a configurable threshold is passed, the kernel can use git to rewind the Tablet to the state at which the currently active anchor was first chosen. In the run on the Brada\v{c} paper this measure is well behaved (Figure~\ref{fig:ps-coarse-progress}): the count of shallowly-closed coarse milestones climbs monotonically from $0$ at the start of formalization to all $\coarsetotal{}$ at the end, pausing only for one extended plateau (cycles~$\coarseplatstart$--$\coarseplatend$) that tripped a StuckMathAudit that produced concrete advice for the reviewer.



\begin{table}[t]
\centering
\caption{Worker outcomes and the kernel's response. A \emph{semantic
delta} is a content change or any non-\textsf{Same} structural update;
no-delta Valid skips verifiers. Stuck/NeedsRestructure roll the
worktree back to the active base before escalating.}
\label{tab:ps-worker}
\footnotesize
\setlength{\tabcolsep}{3.5pt}
\renewcommand{\arraystretch}{1.05}
\begin{tabular}{@{}lll@{}}
\toprule
\textbf{Outcome} & \textbf{Kernel action} & \textbf{Next stage} \\
\midrule
\textsf{Valid}\,(+\,delta) & apply, schedule verifiers & subst.$\to$corr$\to$\dots \\
\textsf{Valid}\,(no delta) & apply, skip verifiers & Reviewer \\
\textsf{Invalid} & retry if attempt $<$ thr.\ ($2$) & Worker \\
 & else escalate w/ context & Reviewer \\
\textsf{Stuck} & rollback; retry-or-escalate & Worker / Reviewer \\
\textsf{NeedsRestructure} & rollback; no retry & Reviewer \\
\bottomrule
\end{tabular}
\end{table}

\begin{table}[t]
\centering
\footnotesize
\setlength{\tabcolsep}{4pt}
\renewcommand{\arraystretch}{1.15}
\begin{tabular}{@{}p{0.585\linewidth}ccc@{}}
\toprule
\textbf{Content the fingerprint hashes} & \textbf{subst.} & \textbf{corr.} & \textbf{sound} \\
\midrule
\multicolumn{4}{@{}l}{\emph{the node's own content}}\\
\quad \LaTeX{} statement & \checkmark & \checkmark & \checkmark \\
\quad \LaTeX{} proof & & & \checkmark \\
\quad \lean{} type-surface (its own signature) & & \checkmark & \\
\quad \lean{} proof body & & & \\[1pt]
\multicolumn{4}{@{}l}{\emph{content of the nodes it depends on}}\\
\quad \lean{} type-surface & & \checkmark & \\
\quad imported definition's \LaTeX{} statement & & \checkmark & \\
\quad cited node's \LaTeX{} statement & & & \checkmark \\
\bottomrule
\end{tabular}
\caption{What each per-node verifier lane's approved fingerprint hashes
(\textsf{subst.}~$=$~substantiveness, \textsf{corr.}~$=$~correspondence,
\textsf{sound}~$=$~soundness). A lane reopens on a node whenever any content
it hashes changes; because a node's \lean{} type-surface includes the types of
everything it imports, editing one node's signature reopens correspondence on
every node that imports it, while reorganizing a \lean{} proof body reopens
nothing.}
\label{tab:ps-fingerprint}
\end{table}

\begin{table}[t]
\centering
\caption{Reviewer directives and the kernel guards that bound them. All
adjudications require the target to be in the just-voted
\texttt{latest\_*\_review} frontier with status \emph{Unknown}.}
\label{tab:ps-directives}
\footnotesize
\setlength{\tabcolsep}{3pt}
\begin{tabular}{@{}p{1.9cm}p{4.9cm}@{}}
\toprule
\textbf{Directive} & \textbf{Key guard} \\
\midrule
\textsc{Continue} (local) & \texttt{authorized\_nodes}\,$=\emptyset$; edits active node only \\
task blocker & worker scope must cover the carrier (Local soundness carve-out for active node) \\
reset blocker & theorem-stating only; current state Fail only \\
restructure & \texttt{authorized\_nodes}\,$\neq\emptyset\subseteq$ scope env.\,$\cap$ anchor cone \\
LastCommit / LastClean & exclusive with partition; LastClean needs a prior clean checkpoint \\
\textsc{Advance\-Phase} / \textsc{Done} & legal only when blockers $=\emptyset$; routes to human gate / Complete \\
\bottomrule
\end{tabular}
\end{table}

\begin{table}[t]
\centering
\caption{The scope ladder: the existing nodes a worker may edit each cycle. The reviewer picks a mode; the kernel bounds the \emph{envelope}, and the reviewer hands the worker an explicit \texttt{authorized\_nodes} subset of it. The active node is a scope anchor, not edit permission. Modes carry the orthogonal closure directives of Table~\ref{tab:ps-directives}.}
\label{tab:ps-scope}
\footnotesize
\setlength{\tabcolsep}{3.5pt}
\renewcommand{\arraystretch}{1.1}
\begin{tabular}{@{}l p{2.7cm} p{2.5cm}@{}}
\toprule
\textbf{Mode} & \textbf{Authorizes (existing nodes)} & \textbf{Notes} \\
\midrule
\multicolumn{3}{@{}l}{\emph{theorem stating}}\\
\quad global       & all present nodes                       & --- \\
\quad targeted     & impact region of the focus             & bidirectional dep-closure \\[1pt]
\multicolumn{3}{@{}l}{\emph{proof formalization}}\\
\quad local        & active node's proof $+$ new helpers     & \texttt{authorized}\,$=\varnothing$ \\
\quad restructure  & enumerated subset of the impact region & may re-sign helpers \\
\quad coarse-restr.\ & as restructure, $+$ protected coarse nodes & only mode that reopens approved correspondence \\[1pt]
\multicolumn{3}{@{}l}{\emph{cleanup}}\\
\quad cleanup      & the cleanup target node                & lint / substitution \\
\bottomrule
\end{tabular}
\end{table}



\tightparagraph{What's left out here.} We omit configurable cadence
constants (audit cooldown/re-audit intervals, retry thresholds), the orphan-cleanup detour scheduled when a delta orphans nodes, and many other implementation details, the authoritative reference for which are the Rust kernel (\texttt{engine.rs}, \texttt{model.rs}) and the \tla{} specification \cite{source}.

\subsection{Trust model}
It is worth stating precisely what \system{} trusts, and for what. We trust the supervisor's deterministic checks---the isolated \lean{} build and the axiom check---to certify that each node's \lean{} proof genuinely supports its \lean{} statement; this is machine-checked and rests on no agent. We trust the human review of the paper targets' semantic closure (Figure~\ref{fig:targetcore}) to certify that the formalization is faithful to those targets: a human ratifies the \lean{} shapes of exactly the declarations the targets' meaning depends on. And we trust the quality of the agents' work---workers, verifier lanes, and the reviewer---only for \emph{progress}; no trust in any agent, the verifier lanes included, is required to trust that a completed formalization is faithful to its targets. Faithfulness and correctness thus reduce to the first two; the agents determine only whether \system{} reaches a completed state, not whether that state is sound.

\section{Relation to Prior Work}
\label{sec:related}

Work on automated and neural theorem proving has produced systems that search for proof terms~\cite{gptf}, integrate learned models with automated provers~\cite{thor}, repair whole proofs with language models~\cite{baldur}, or retrieve premises to prove theorems inside proof assistants~\cite{leandojo}. Benchmarks such as miniF2F~\cite{minif2f} have made formal proof generation more measurable, and Lean with its mathematical library has made large-scale formalized mathematics practically accessible. A parallel line studies autoformalization itself. One strand frames it as a long-term, learning-driven route to general mathematical reasoning, in which a system bootstraps the ability to read and formalize mathematics largely from data~\cite{szegedy}; another shows empirically that few-shot LLMs can translate individual competition-problem \emph{statements} into formal specifications, the resulting data then improving a neural prover~\cite{autoformalization}. \system{} departs from both: it adds no task-specific training---its specialization is the meaning-of-rigor motivated process semantics, not a learned formalization model---and its unit of work is a whole paper's proof, decomposed and \lean{}-closed end to end, rather than an isolated statement.

\system{} is complementary. It does not replace proof search, tactic prediction, or statement translation. It supplies a process layer around them. Its target is the gap between a paper proof and a final formal artifact: the long interval in which the system must decide whether partially formalized statements, lemmas, and proof sketches are genuine progress. In that sense, \system{} is closer to a workflow semantics for mathematical refinement than to a standalone prover.  It should be noted also that the immediate aim of \system{} is excellence at the task ``autoformalize a given paper'' with no guidance, which is quite different from the task of building useful infrastructure~\cite{mathlib}, where every choice made is essential for further utility. Needless to say, the feasibility of the task \system{} aims for excellence at is tied to the availability of suitable infrastructure\footnote{We do take the position that the development of autoformalization should not wait until such infrastructure is ``complete''; we believe that many hard papers can be autoformalized today, and that developing excellent systems for this task both provides real value and a stronger impetus for infrastructure development.}, and in this way emphasizes, rather than obviates, the value of such infrastructure\footnote{Kontorovich has discussed the problem of closing the loop from autoformalization to infrastructure-building under the name \emph{auto-canonization}~\cite{kontorovich_canon}.}.

The closest conceptual relatives are human-in-the-loop formalization environments and agentic proof assistants. Kontorovich~\cite{kontorovich} describes a closely related ``quasi-autoformalization'' workflow, in which a decomposer, translator, solver, and conductor help turn mathematical prose into Lean, with human intervention available when useful. Trellis shares this basic motivation but, at least in its form today with a focus on end-to-end formalization, assigns complete semantic authority to a deterministic kernel rather than to an agentic conductor. The goal is to make progress a machine-checkable property of the tablet state---through substantiveness, correspondence, soundness, and Lean closure. It also provides a way of optimizing and benchmarking the fully automatic part of autoformalization, even when imagining implementations that keep a human in the driver's seat.

\tightparagraph{Acknowledgement:} We thank Alex Kontorovich for helpful conversations which improved an earlier draft of this document.

\end{document}